\documentclass[conference]{IEEEtran}
\IEEEoverridecommandlockouts
\usepackage{cite}
\usepackage{amsmath,amssymb,amsfonts,multirow,multicol,subfigure}
\usepackage[table]{xcolor}
\usepackage{algorithm,algorithmic}
\usepackage{graphicx}
\usepackage{textcomp}
\usepackage{xcolor}
\usepackage{balance}
\usepackage{hyperref}
\def\BibTeX{{\rm B\kern-.05em{\sc i\kern-.025em b}\kern-.08em
    T\kern-.1667em\lower.7ex\hbox{E}\kern-.125emX}}
\begin{document}

\title{A Simple Evolutionary Algorithm for Multi-modal Multi-objective Optimization
\thanks{The first and third authors would like to acknowledge the support from Australian Research Council Discovery Project DP220101649. }
}

\author{\IEEEauthorblockN{Tapabrata Ray}
\IEEEauthorblockA{\textit{School of Engineering and IT} \\
\textit{The University of New South Wales}\\
Canberra, Australia \\
t.ray@adfa.edu.au}
\and
\IEEEauthorblockN{Mohammad Mohiuddin Mamun}
\IEEEauthorblockA{\textit{School of Engineering and IT} \\
\textit{The University of New South Wales}\\
Canberra, Australia \\
m.mamun@student.adfa.edu.au}
\and
\IEEEauthorblockN{Hemant Kumar Singh}
\IEEEauthorblockA{\textit{School of Engineering and IT} \\
\textit{The University of New South Wales}\\
Canberra, Australia \\
h.singh@adfa.edu.au}
}

\maketitle

\begin{abstract}
In solving multi-modal, multi-objective optimization problems~(MMOPs), the objective is not only to find a good representation of the Pareto-optimal front~(PF) in the objective space but also to find all equivalent Pareto-optimal subsets~(PSS) in the variable space. Such problems are practically relevant when a decision maker~(DM) is interested in identifying alternative designs with similar performance. There has been significant research interest in recent years to develop efficient algorithms to deal with MMOPs. However, the existing algorithms still require prohibitive number of function evaluations~(often in several thousands) to deal with problems involving as low as two objectives and two variables. The algorithms are typically embedded with sophisticated, customized mechanisms that require additional parameters to manage the diversity and convergence in the variable and the objective spaces. In this paper, we introduce a steady-state evolutionary algorithm for solving MMOPs, with a simple design and no additional user-defined parameters that need tuning compared to a standard EA. We report its performance on $21$ MMOPs from various test suites that are widely used for benchmarking using a low computational budget of $1000$ function evaluations. The performance of the proposed algorithm is compared with six state-of-the-art algorithms~(MO\_Ring\_PSO\_SCD, DN-NSGAII, TriMOEA-TA\&R, CPDEA, MMOEA/DC and MMEA-WI). The proposed algorithm exhibits significantly better performance than the above algorithms based on the established metrics including IGDX, PSP and IGD. We hope this study would encourage design of simple, efficient and generalized algorithms to improve their uptake for practical applications.{\footnote{This is the accepted version of the paper at the IEEE Congress on Evolutionary Computation (CEC) 2022. Published proceedings version available at \url{https://ieeexplore.ieee.org/document/9870274/}}} 
\end{abstract}

\section{Background}
There has been significant research effort directed towards development of algorithms to deal with multi-modal multi-objective optimization problems~(MMOPs) in recent years, as evident from a selected bibliographic review in Table~\ref{tab:Algo table}. We summarize key observations from the list before outlining the motivation of this work.
\begin{itemize}
    \item While this list is by no means comprehensive, one can observe that attempts to solve MMOPs have been reported as early as 2005. While the number of studies reported on MMOPs have been relatively low in the evolutionary multi-objective optimization~(EMO) domain, past few years have seen a significant rise in such numbers. From 2018 onwards in particular, there has been a proliferation of algorithms in this domain. For a recent review of the existing works, the readers are referred to the survey paper~\cite{tanabe2019review}. The survey also highlights the difficulty in accounting for all such works since many of them do not explicitly mention multi-modality even though the underlying problem is often multi-modal in nature.
    \item Most of the studies deal with problems involving $2-3$ variables and $2-3$ objectives. Exceptions to this include up to $10$ variables in DESS~\cite{peng2021diversity} and $15$ variables in NIMMO~\cite{tanabe2019niching} and MMEA-WI~\cite{li2021weighted} papers.
    \item Even when dealing with MMOPs involving $2-3$ variables and $2-3$ objectives, the population size and the number of allowable function evaluations are significantly high in the contemporary studies. Population sizes typically range between $200$ and $1000$ and the number of function evaluations range between $10,000$ and $500,000$. Utilizing these large numbers may be untenable for the applications where the evaluation of each candidate solution is computationally expensive~(e.g. numerical simulation or physical experiments in-loop). Therefore, development of algorithms that could solve MMOPs with much fewer function evaluations is needed and would add significant value to this field.
    \item The performance of the algorithms has been assessed using problems from relatively new test suites such as MMF~\cite{yue2017multiobjective}, MMMOP~\cite{liu2018multimodal}, IDMP~\cite{liu2019handling} etc. These test suites have been introduced in recent years to capture various characteristics of the Pareto sets/subsets to encourage development of algorithms to address them. Among the more seminal test problems, SYM-PART~\cite{rudolph2007capabilities} and Omni-test~\cite{deb2005omni} are also frequently used in a number of works. 
    \item In terms of the metrics for performance assessment, the studies typically use IGDX~\cite{zhou2009approximating}, PSP~\cite{yue2017multiobjective}, IGD~\cite{riquelme2015performance} or IGDM~\cite{liu2018multimodal}. Among these, IGD quantifies the performance only in the objective space. While it is quite commonly used in the EMO field for benchmarking generic multi-objective evolutionary algorithms~(MOEAs), it is insufficient in itself in the context of MMOPs since the task at hand requires coverage in PS along with PF. The other metrics mentioned above have been developed more recently to facilitate systematic, quantitative benchmarking of the algorithms aimed at solving MMOPs. These metrics take into account the the solutions obtained in the variable space rather than objective space. 
    \item A closer look at the prominent algorithms makes it apparent that most of them achieve competitive performance though the use of a number of bespoke mechanisms which often involve additional parameters and in turn lead to increased complexity. For such complex algorithms, it is often difficult to conclusively understand the effects/contributions of the underlying schemes and their interactions. Moreover, the need for tuning the parameters might prove to be an impediment in adoption of the algorithms more widely in practical applications. 
\end{itemize}

\begin{table*}[!htb]\scriptsize
\centering
\caption{Experimental settings used in contemporary studies. Notation `$M$' refers to the number of objectives and `$D$' refers for the number of variables.}
\label{tab:Algo table}
\renewcommand{\arraystretch}{1.1}
\tabcolsep 1mm
\begin{tabular}{|c|c|c|c|c|c|c|c|}
\hline
\multicolumn{1}{|c|}{\textbf{Algorithm Name~[Reference]}} & \textbf{Year} & \textbf{$M$/$D$} & \textbf{Population Size}                                       & \textbf{Function Evaluations}                       & \textbf{Test Problems}                                   \\ \hline

\multicolumn{1}{|c|}{\multirow{2}{*}{Omni-Optimizer~\cite{deb2005omni}}}           & \multirow{2}{*}{2005}& \multirow{2}{*}{2 / 2}           & \multirow{2}{*}{1000}     & \multirow{2}{*}{500000}  & \multirow{2}{*}{SYM-PART} \\ 
&&&&& \\\hline

\multicolumn{1}{|c|}{\multirow{2}{*}{DBSCAN~\cite{kramer2010dbscan}}}           & \multirow{2}{*}{2010}  & \multirow{2}{*}{2 / 2}  & \multirow{2}{*}{300}     & \multirow{2}{*}{20000}  & \multirow{2}{*}{SYM-PART, Two-On-One} \\ 
&&&&& \\\hline

\multicolumn{1}{|c|}{\multirow{2}{*}{DN-NSGAII  ~\cite{liang2016multimodal}}}           & \multirow{2}{*}{2016}  & \multirow{2}{*}{2 / 2}         & \multirow{2}{*}{800}     & \multirow{2}{*}{80000}  & \multirow{2}{*}{SYM-PART, SS-UF} \\ 
&&&&& \\\hline

\multicolumn{1}{|c|}{\multirow{2}{*}{MO\_Ring\_PSO\_SCD~\cite{yue2017multiobjective}}}           & \multirow{2}{*}{2017} & \multirow{2}{*}{2 / 2-3}          & \multirow{2}{*}{800}     & \multirow{2}{*}{80000}  & \multirow{2}{*}{MMF, SYM-PART, Omni-test} \\ 
&&&&& \\\hline

\multicolumn{1}{|c|}{\multirow{2}{*}{TriMOEA-TA\&R~\cite{liu2018multimodal}}}           & \multirow{2}{*}{2018}  & \multirow{2}{*}{2-3 / 2-7}        & \multirow{2}{*}{800}     & \multirow{2}{*}{80000}   & \multirow{2}{*}{MMMOP, MMF} \\ 
&&&&& \\\hline

\multicolumn{1}{|c|}{\multirow{2}{*}{MOEA/D-D-AD~\cite{tanabe2018decomposition}}}           & \multirow{2}{*}{2018}  & \multirow{2}{*}{2 / 2,5}          & \multirow{2}{*}{100}     & \multirow{2}{*}{30000} & {SYM-PART, SS-UF, } \\ 
&&&&&{Two-On-One, Omni-test} \\\hline

\multicolumn{1}{|c|}{\multirow{2}{*}{DNEA~\cite{liu2018double}}}           & \multirow{2}{*}{2018} & \multirow{2}{*}{2 / 2}          & \multirow{2}{*}{210}     & \multirow{2}{*}{63000}  & \multirow{2}{*}{Polygon} \\ 
&&&&& \\\hline

\multicolumn{1}{|c|}{\multirow{2}{*}{MOEA/D-Variant~\cite{hu2018incorporation}}}           & \multirow{2}{*}{2018} & \multirow{2}{*}{2 / 2}         & \multirow{2}{*}{1120}     & \multirow{2}{*}{89600}   & \multirow{2}{*}{MMDMP} \\ 
&&&&& \\\hline

\multicolumn{1}{|c|}{\multirow{2}{*}{NIMMO~\cite{tanabe2019niching}}}           & \multirow{2}{*}{2019} & \multirow{2}{*}{(2,3,5,8-10,15) /}        & {200($M$=2), 210($M$=3,5,9), 156($M$=8)}      & \multirow{2}{*}{10000}  & {MMF, SYM-PART, Omni-test, } \\ 
&&{(2,5)}   & {230($M$=10), 135($M$=15)} &&{Two-On-One, Polygon}  \\\hline

\multicolumn{1}{|c|}{\multirow{2}{*}{MMODE~\cite{liang2019multimodal}}}           & \multirow{2}{*}{2019}  & \multirow{2}{*}{2 / 2}         & \multirow{2}{*}{800}     & \multirow{2}{*}{80000}  & \multirow{2}{*}{MMF, Omni-test} \\ 
&&&&& \\\hline

\multicolumn{1}{|c|}{\multirow{2}{*}{DE-RLFR~\cite{li2019differential}}}           & \multirow{2}{*}{2019}  & \multirow{2}{*}{2 / 2-3}         & \multirow{2}{*}{800}     & \multirow{2}{*}{80000}  & \multirow{2}{*}{MMF, SYM-PART, Omni-test} \\ 
&&&&& \\\hline

\multicolumn{1}{|c|}{\multirow{2}{*}{MMO-CLRPSO~\cite{zhang2019cluster}}}           & \multirow{2}{*}{2019}  & \multirow{2}{*}{2 / 2-3}         & \multirow{2}{*}{800}     & \multirow{2}{*}{80000}  & \multirow{2}{*}{MMF, SYM-PART, Omni-test} \\ 
&&&&& \\\hline

\multicolumn{1}{|c|}{\multirow{2}{*}{CPDEA~\cite{liu2019handling}}}           & \multirow{2}{*}{2019}  & \multirow{2}{*}{2 / 2-4}        & {60($M$=2), 120($M$=3)}     & {18000($M$=2), 36000($M$=3)}   & \multirow{2}{*}{IDMP} \\ 
&&& 240($M$=4) & 72000($M$=4)  & \\\hline

\multicolumn{1}{|c|}{\multirow{2}{*}{MMODE\_CSCD~\cite{liang2021clustering}}}           & \multirow{2}{*}{2021}  & \multirow{2}{*}{2-3 / 2-3}         & \multirow{2}{*}{200}     & \multirow{2}{*}{20000}  & \multirow{2}{*}{MMF, SYM-PART, Omni-test} \\ 
&&&&& \\\hline

\multicolumn{1}{|c|}{\multirow{2}{*}{MMODE\_ICD~\cite{yue2021differential}}}           & \multirow{2}{*}{2021}  & \multirow{2}{*}{2-3 / 2-3}         & \multirow{2}{*}{200(D=2), 300(D=3)}     & \multirow{2}{*}{10000(D=2), 15000(D=3)}  & \multirow{2}{*}{MMF, SYM-PART, Omni-test} \\ 
&&&&& \\\hline

\multicolumn{1}{|c|}{\multirow{2}{*}{MMO\_SO\_QPSO~\cite{li2021handling}}}           & \multirow{2}{*}{2021}  & \multirow{2}{*}{2-3 / 2-3}         & \multirow{2}{*}{200(D=2), 300(D=3)}     & \multirow{2}{*}{10000(D=2), 15000(D=3)}  & {MMF, SYM-PART, Omni-test,}  \\ 
&&&&&{MMMOP} \\\hline

\multicolumn{1}{|c|}{\multirow{2}{*}{MMOEA/DC~\cite{lin2020multimodal}}}           & \multirow{2}{*}{2021}  & \multirow{2}{*}{2-3 / 2-3}         & \multirow{2}{*}{200(D=2), 300(D=3)}     & \multirow{2}{*}{10000(D=2), 15000(D=3)}  & \multirow{2}{*}{MMF, SYM-PART, Omni-test}  \\ 
&&&&& \\\hline

\multicolumn{1}{|c|}{\multirow{2}{*}{MMEA-WI~\cite{li2021weighted}}}           & \multirow{2}{*}{2021}  & {(2,3-5,8-10,15) /}         & {200($M$=2), 210($M$=3-5,9), 156($M$=8)}     & {10000($M$=2), 12000($M$=3)}  & \multirow{2}{*}{MMF, IDMP, Polygon} \\ 
&& {(2-4,10,100)}& {230($M$=10), 135($M$=15)} & {15000($M$=4), 20000($M>$4)}  & \\\hline

\multicolumn{1}{|c|}{\multirow{2}{*}{DN-MMOES~\cite{zhang2021two}}}           & \multirow{2}{*}{2021} & \multirow{2}{*}{2-4 / 2-4}          & \multirow{2}{*}{200}     & \multirow{2}{*}{200000}  & \multirow{2}{*}{MMF, IDMP, SYM-PART, Omni-test} \\ 
&&&&& \\\hline

\multicolumn{1}{|c|}{\multirow{2}{*}{MMO-MOES~\cite{zhang2021multi}}}           & \multirow{2}{*}{2021}   & \multirow{2}{*}{2-3 / 2-3}        & \multirow{2}{*}{400}     & \multirow{2}{*}{120000}  & \multirow{2}{*}{MMF, SYM-PART, Omni-test} \\ 
&&&&& \\\hline

\multicolumn{1}{|c|}{\multirow{2}{*}{DESS~\cite{peng2021diversity}}}           & \multirow{2}{*}{2021} & \multirow{2}{*}{2-3 / 2,4,6,10}          & \multirow{2}{*}{300}     & \multirow{2}{*}{50000}  & \multirow{2}{*}{MMF, SYM-PART, Omni-test, Polygon } \\ 
&&&&& \\\hline

\end{tabular}
\end{table*}

With this backdrop, in this paper, we first present the scope and intent of this study.
\begin{itemize}
\item We introduce a simple algorithm for solving MMOPs, without the use of additional parameters than those ordinarily required in an MOEA.
\item The scope is limited to the algorithms that are listed to have capability to deal with MMOPs as in PlatEMO framework~(Version 3.3, released 2021/08/14)~\cite{tian2017platemo} and few others for which we obtained the source codes from the authors.
\item The analysis is based on $21$ test problems from MMF, MMMOP, SYM-PART and Omni-test suites that are commonly used in contemporary studies assessed using established metrics such as IGDX, PSP and IGD.
\item The intent is to observe if such an algorithm can deliver competitive performance using a smaller computational budget of $1000$ function evaluations. The development is also based on an assumption that the computational costs associated with all the algorithmic steps are insignificant when compared to the cost of evaluating the solutions, that is common for computationally expensive optimization problems. We would also like to point out that while surrogate-assisted optimization approaches are commonly used to deal with such classes of problems, we would want to improve the performance of the baseline optimization algorithm itself before embedding any forms of surrogates.
\end{itemize}

The proposed algorithm is presented in Section~\ref{sec:PA}. The numerical experiments are detailed in Section~\ref{sec:NR}, followed by concluding remarks in Section~\ref{sec:SC}. 

\section{Proposed Algorithm}
\label{sec:PA}
The proposed algorithm for solving multi-modal, multi-objective optimization is referred to as MOMO. MOMO operates in a \emph{steady-state} framework, implying that only one solution is selected for evaluation in each generation. The choice of steady-state paradigm is to (a) make the algorithm tenable to cases where the computational budget is extremely limited, and (b) make the algorithm suitable for cases where the evaluations cannot be parallelized. This can happen when, e.g., only once software license or physical experimental setup is available to conduct the evaluation. 

The pseudo-code of MOMO is presented in Algorithm~\ref{algo:base}. The algorithm starts with an initial population of $N$ solutions generated using uniform random distribution within the variable bounds. These solutions are evaluated and two parents are selected and recombined using simulated binary crossover~(SBX)~\cite{deb2002fast} and polynomial mutation~(PM)~\cite{deb2002fast} to create two offspring, of which a random one is evaluated. From the resulting $N+1$ solutions~($N$ parents and $1$ offspring), the best $N$ solutions are carried forward as the surviving population for the next generation by eliminating one solution. These steps are repeated until computational budget is exhausted. It is a simple steady-state algorithm and the only key elements that need elaboration are (a) the scheme for identifying two parents for recombination and (b) the scheme for environmental selection of $N$ solutions from $N+1$ solutions.

\begin{algorithm}[!ht]
	\caption{Proposed Algorithm: MOMO}
	\parbox[l]{\linewidth}{
	\algorithmicrequire\hspace{1mm}$NFE_{max}$\hspace{1mm}(maximum number of function evaluations), $N$\hspace{1mm}(population size), probability of SBX crossover\hspace{1mm}$P_c$, distribution index of SBX crossover\hspace{1mm}$\eta_c$, probability of polynomial mutation\hspace{1mm}$P_m$, distribution index of polynomial mutation\hspace{1mm}$\eta_m$.}
	\scalebox{0.94}{
	\begin{minipage}[H]{\linewidth}
	\flushleft
	\begin{algorithmic}[1]
    \STATE Set archive $\mathcal{A} = \emptyset$, $NFE$ = 0 \COMMENT{Initial generation}
    \STATE $P \leftarrow$ Initialize population \COMMENT{Uniform random distribution within the variable bounds} 
    \STATE $P \leftarrow$ Evaluate $P$
    \STATE $NFE \leftarrow NFE+N$ \COMMENT{Update NFE}
    \STATE $\mathcal{A} \leftarrow \mathcal{A} \cup P$ \COMMENT{Add evaluated solutions to archive}
	 \WHILE{$(NFE\le NFE_{max})$}
	    \STATE $PP \leftarrow $  Select two solutions from $P$ for recombination \COMMENT{Details in Section~\ref{subsec:recomb}}
	    \STATE $C \leftarrow $ Recombine the two solutions in $PP$ using SBX and PM and select the first offspring.
	    \STATE Evaluate~$C$. 
	    \STATE $NFE \leftarrow NFE+1$ \COMMENT{Update NFE}
	    \STATE $\mathcal{A} \leftarrow \mathcal{A} \cup C$.
	    \STATE $P \leftarrow$ Environmental selection ($P \cup C$) \COMMENT{Details in Section~\ref{subsec:env}}
	 \ENDWHILE			
	\end{algorithmic}
	\vspace{1mm}
	\vspace{1mm}
	\end{minipage}}
	\label{algo:base}
\end{algorithm}

\subsection{Parent selection for recombination}
\label{subsec:recomb}

Given the $N$ solutions in a population, we first identify their ranks using non-dominated sorting~\cite{deb2002fast}. Then, to select two parents for recombination, we partition $N$ solutions into ``optimal clusters'' in the variable space using Silhouette clustering index~\cite{rousseeuw1987silhouettes}. For this, the variables are normalized in the range [0,1] linearly using the minimum and maximum variable values that occur among these solutions and cluster them using $k-$means~(MATLAB default implementation is used without any customization/tuning). The number of clusters is determined by sequentially trying $k = 2, 3, \dots, K$ until for $k=K$ any resulting cluster consists of only a single solution. It is important to take note that the above process does not require any user input. Then, the Silhouette index is calculated for the each of these $k \in \{2,3,\ldots K\}$ values, and the optimal number of clusters $k^*$ is determined as the one with the best~(highest) Silhouette index. 

It is important to take note that cluster center initialization in $k-$means is heuristic and stochastic. When attempting to use a value of $k$, the process might yield different partitions even for the same data set depending on the initial centers. Therefore, instead of using the instantaneous optimal $k^*$ determined using Silhouette index, we use the \emph{running average}~(nearest ceiling integer) of the optimal $\bar{k}^*$, henceforth referred as \emph{stabilized} number of clusters, for  clustering.

After partitioning $N$ solutions into stabilized number of clusters~($\bar{k}^*$), we pick two clusters that have the smallest numbers of solutions. From each of these clusters, we select one best non-domination ranked solution as a parent for recombination. In case there are multiple solutions with the same~(best) non-domination rank within a cluster, one is randomly selected amongst them as a parent. 

As evident from above, the rationale behind the design of this parent selection scheme is to (a) automatically identify how many PSS are likely, (b) promote diversity by choosing parents from the clusters that currently have relatively few solutions, and (c) promote convergence by choosing parents that have the highest non-domination rank in those clusters.  

\subsection{Environmental selection}
\label{subsec:env}

For environmental selection, we need to eliminate $1$ solution from $N+1$ solutions. The worst non-domination ranked solution from the largest cluster is eliminated in the proposed approach. The rationale behind this scheme is that the largest cluster has the most number of representative solutions to cover the respective PSS, and hence eliminating one from this cluster would least adversely impact the diversity of the overall population. At the same time, choice of eliminating the worst ranked~(in terms of non-dominance) from this cluster will work in favour of convergence. 

In case there are multiple solutions with the same~(worst) rank, a solution is randomly selected amongst them for elimination. The process inherently tries to maintain a population with an equal number of solutions in each PSS approximation. 

While one can replace the random selections~(multiple solutions with the same best rank or same worst rank) in the above components with more sophisticated selection schemes to further improve diversity in variable or objective spaces, we have refrained from doing so to maintain simplicity in the current version. 

\subsection{Proof-of-concept results}

\begin{figure}[!ht]
\centering    
\subfigure[]{\label{fig:Numfig-X-25}\includegraphics[width=0.23\textwidth]{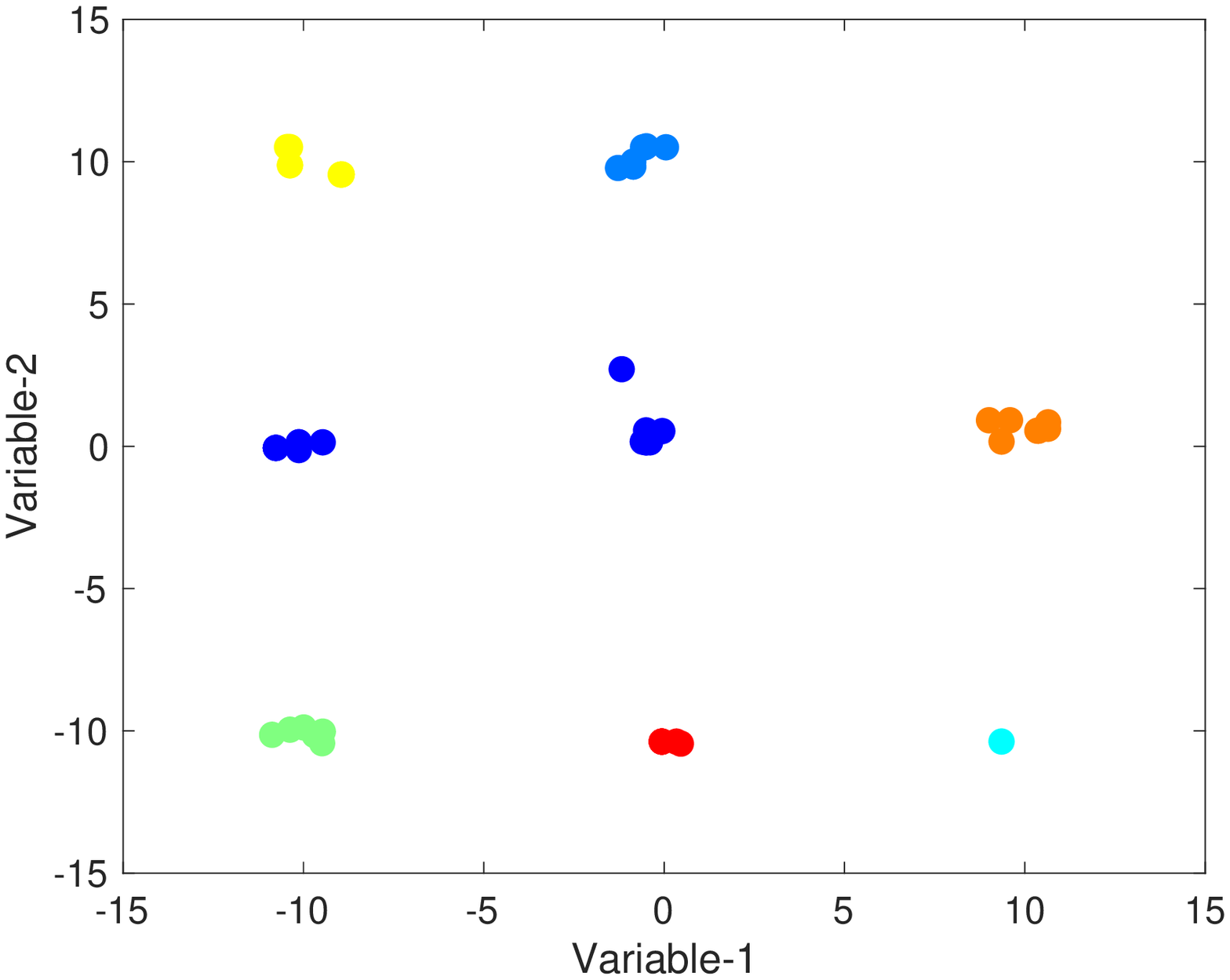}} \quad 
\subfigure[]{\label{fig:Numfig-F-25}\includegraphics[width=0.23\textwidth]{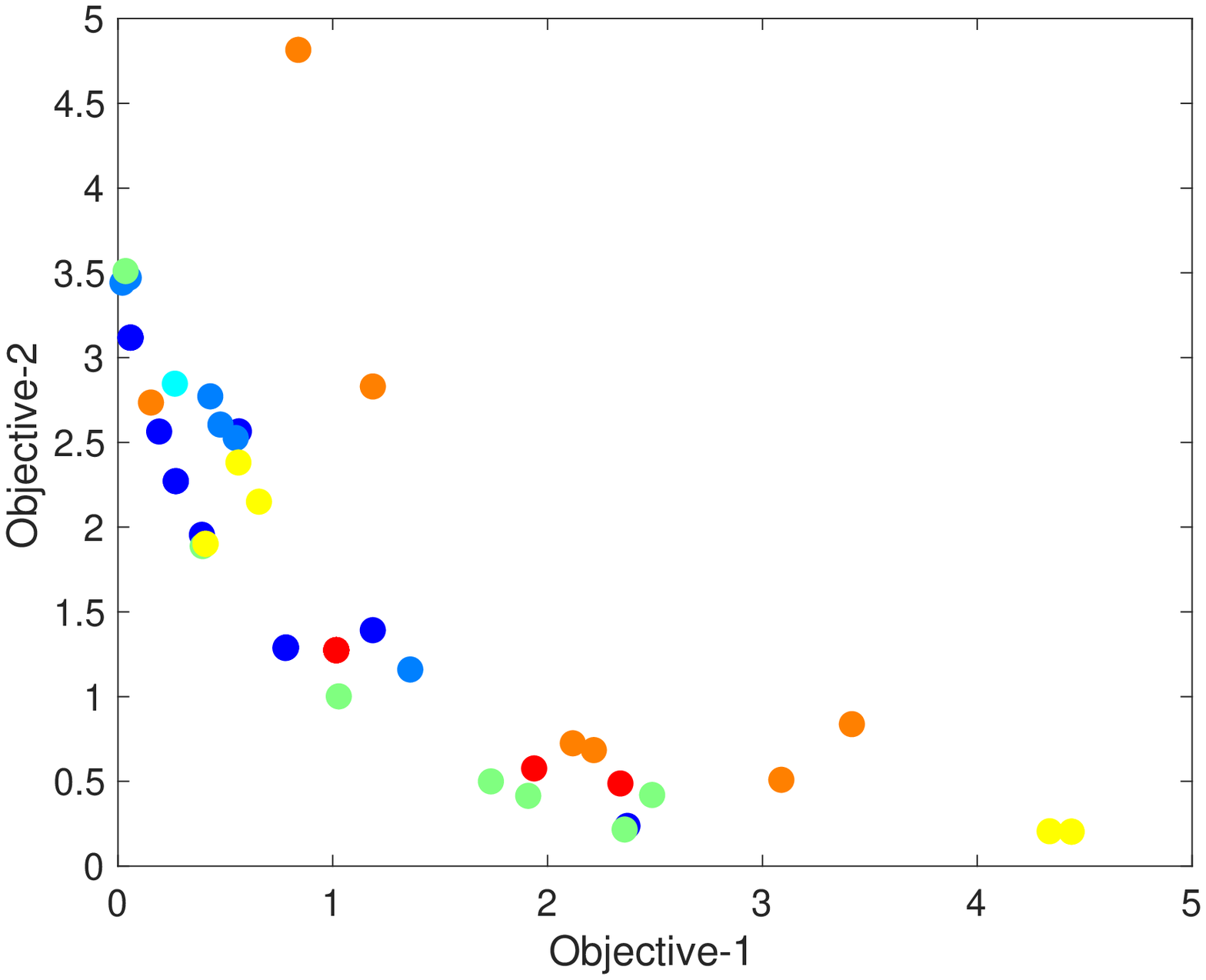}} \\
\subfigure[]{\label{fig:Numfig-X-50}\includegraphics[width=0.23\textwidth]{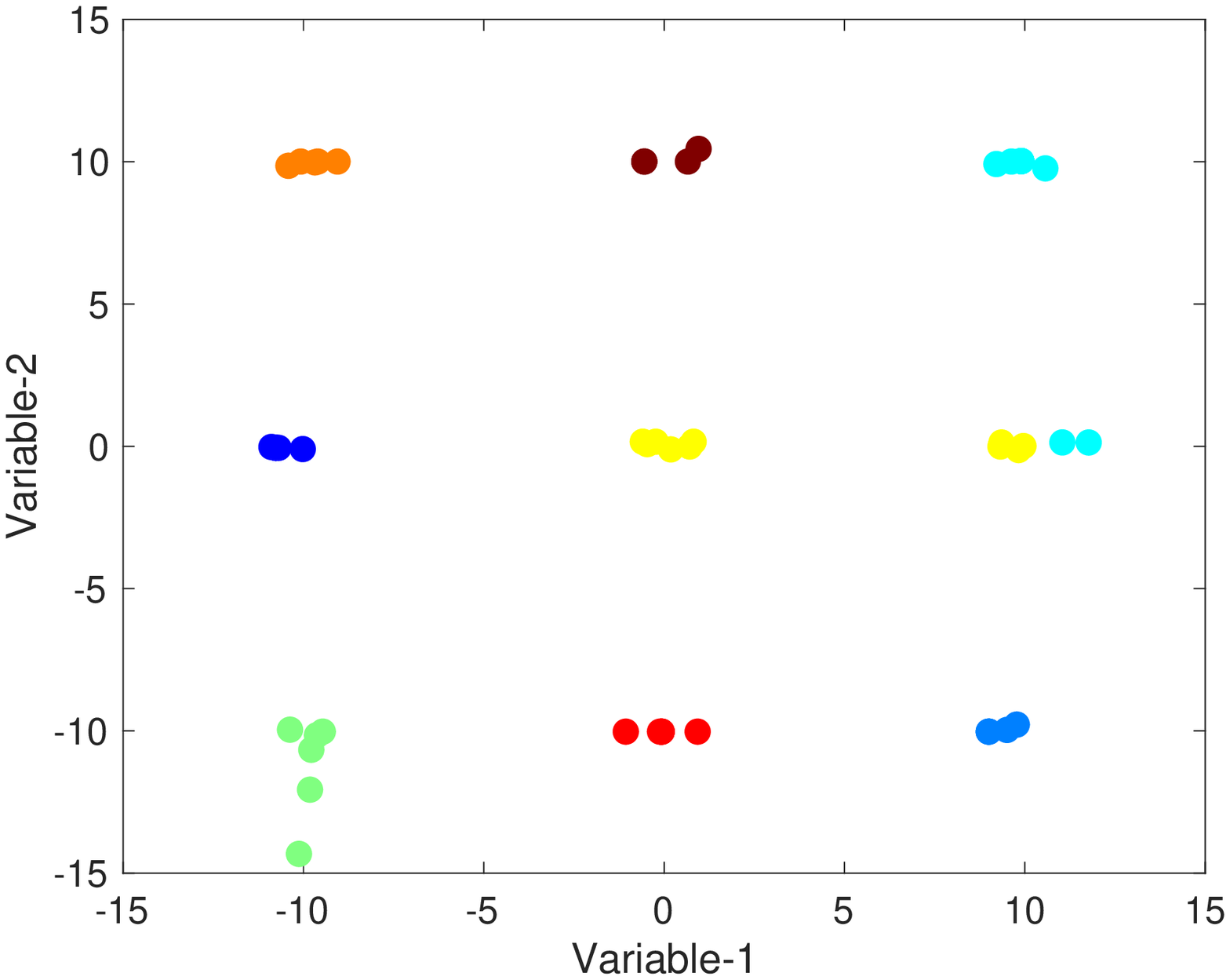}} \quad 
\subfigure[]{\label{fig:Numfig-F-50}\includegraphics[width=0.23\textwidth]{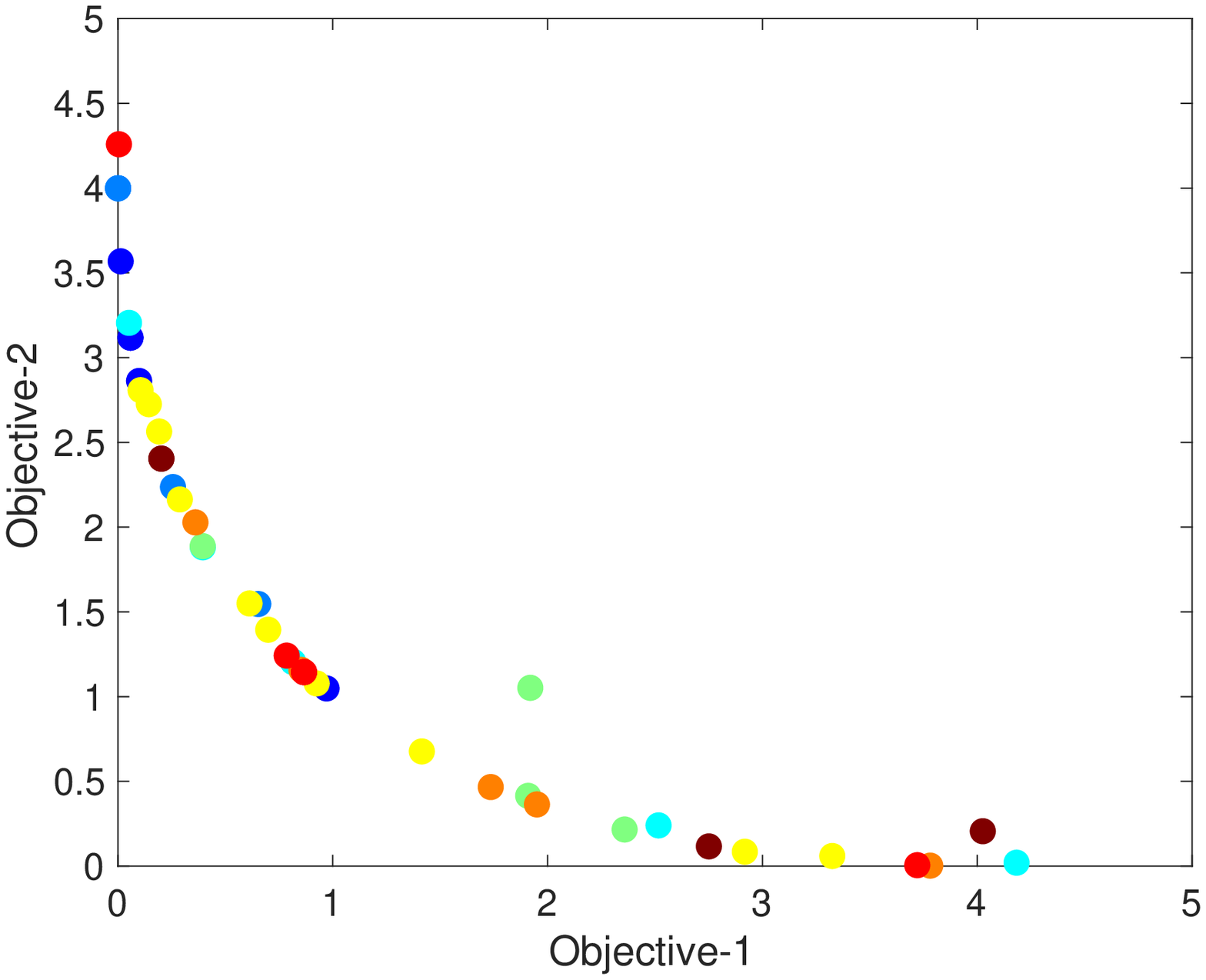}} \\
\subfigure[]{\label{fig:Numfig-X-75}\includegraphics[width=0.23\textwidth]{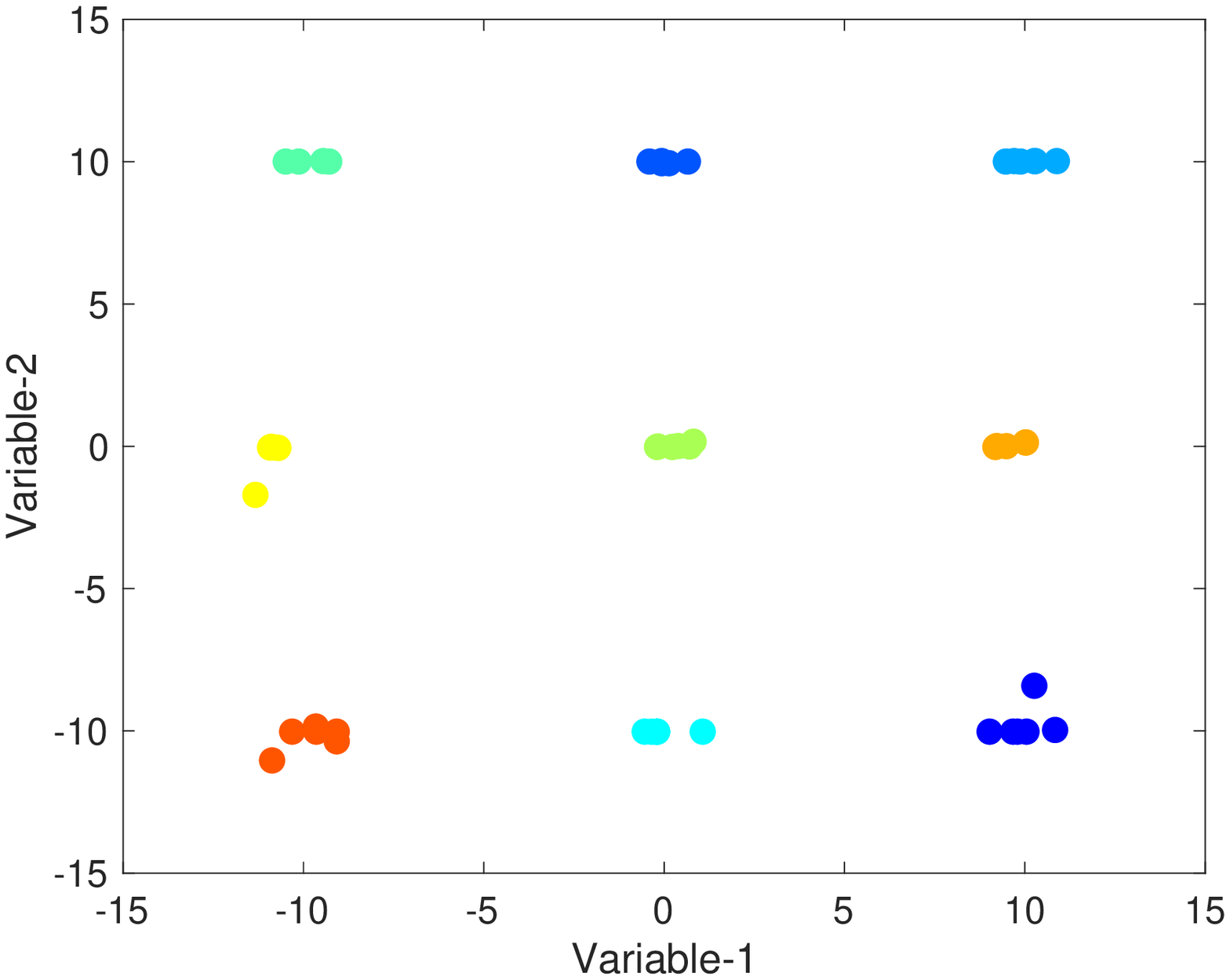}} \quad 
\subfigure[]{\label{fig:Numfig-F-75}\includegraphics[width=0.23\textwidth]{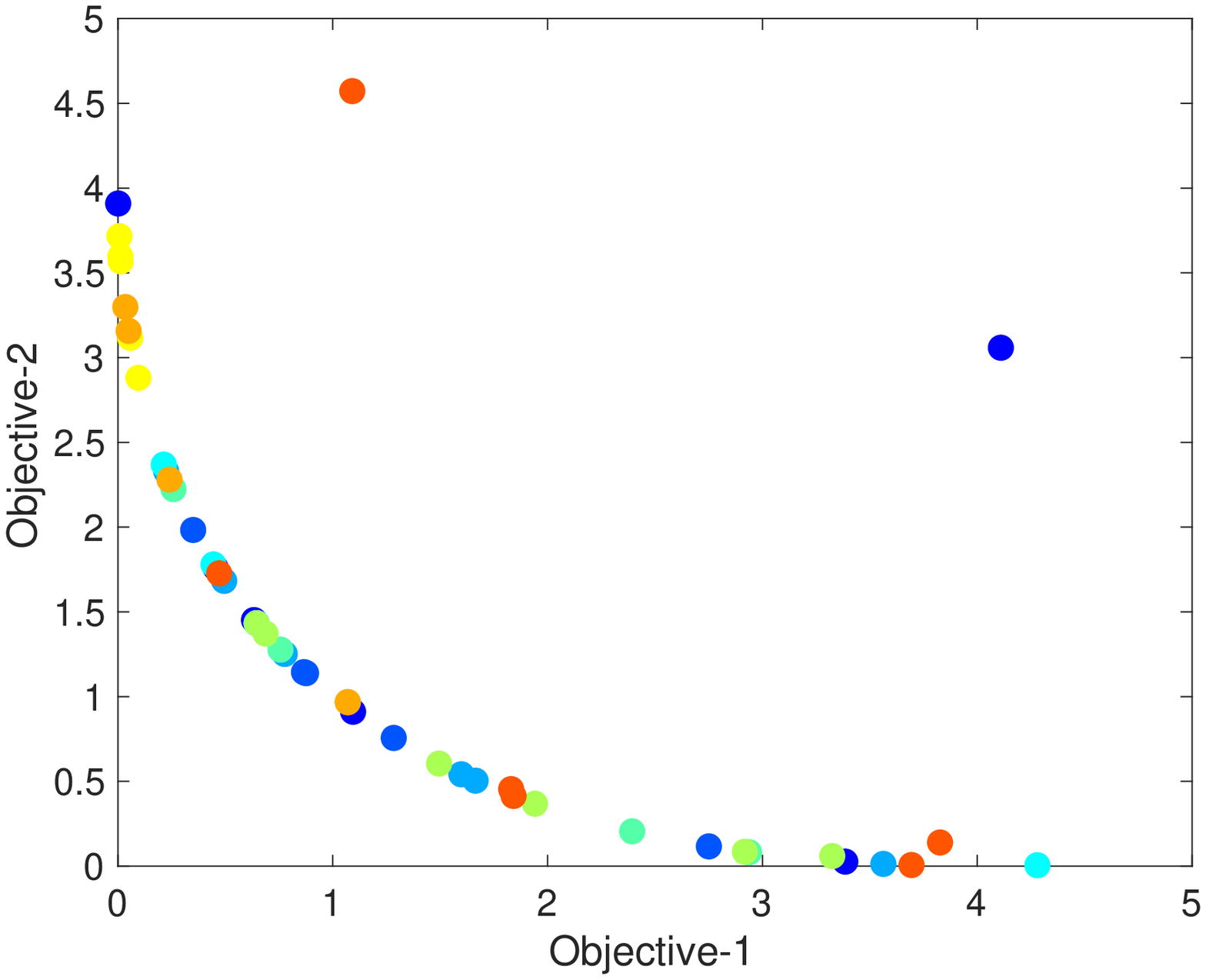}} \\
\subfigure[]{\label{fig:Numfig-X-100}\includegraphics[width=0.23\textwidth]{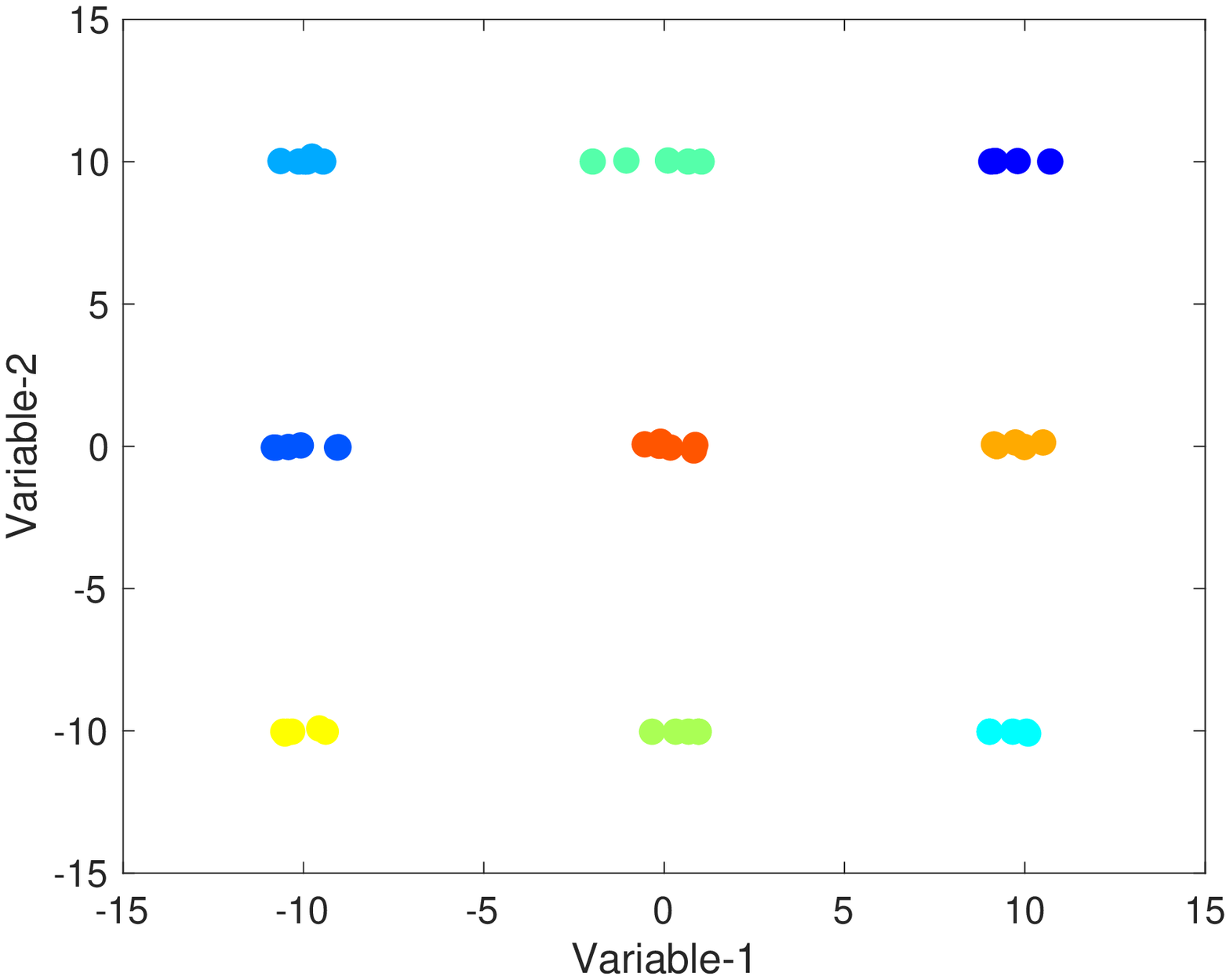}} \quad 
\subfigure[]{\label{fig:Numfig-F-100}\includegraphics[width=0.23\textwidth]{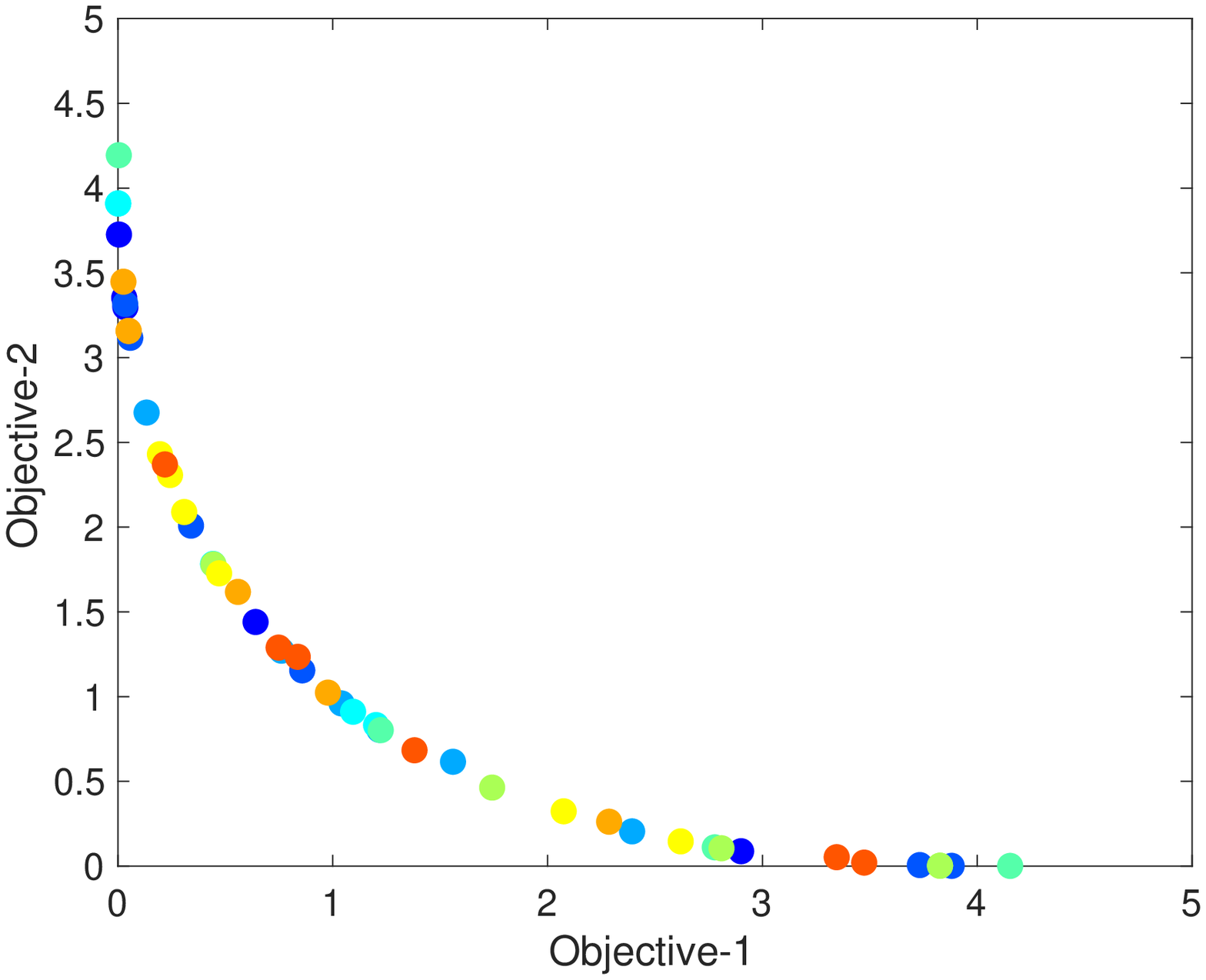}} \\
\caption{The distribution of the population in variable and objective spaces at (from the top, row-wise) $25\%$, $50\%$, $75\%$ and $100\%$ computational budget for SYM-PART Simple problem. Each cluster is marked with a different color.}
\label{fig:NumFig-XF}
\end{figure}

\begin{figure}[!ht]
\centering    
\subfigure[]{\label{fig:Numfig-1}\includegraphics[width=0.36\textwidth]{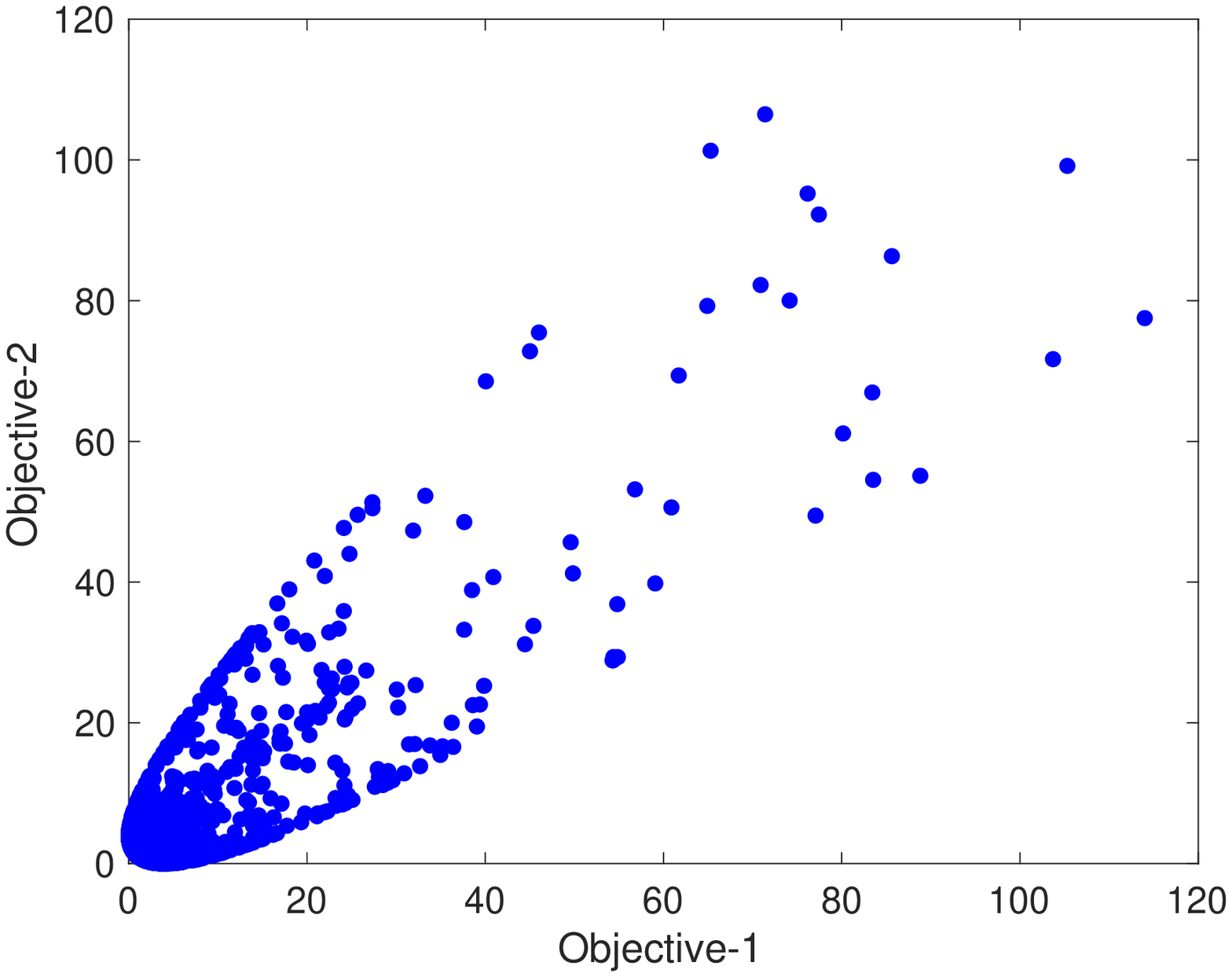}} \\
\subfigure[]{\label{fig:Numfig-2}\includegraphics[width=0.36\textwidth]{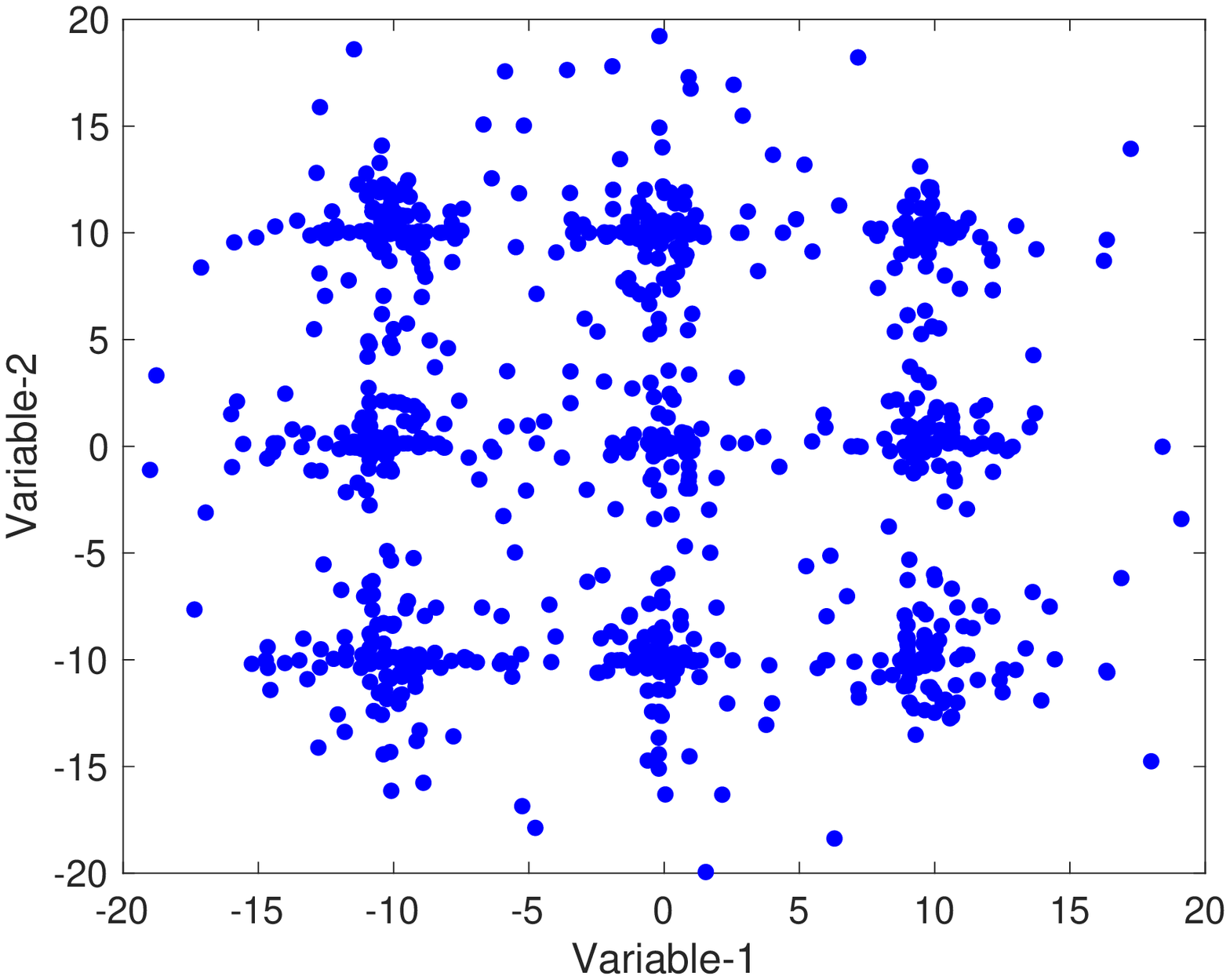}} \\
\subfigure[]{\label{fig:Numfig-3}\includegraphics[width=0.36\textwidth]{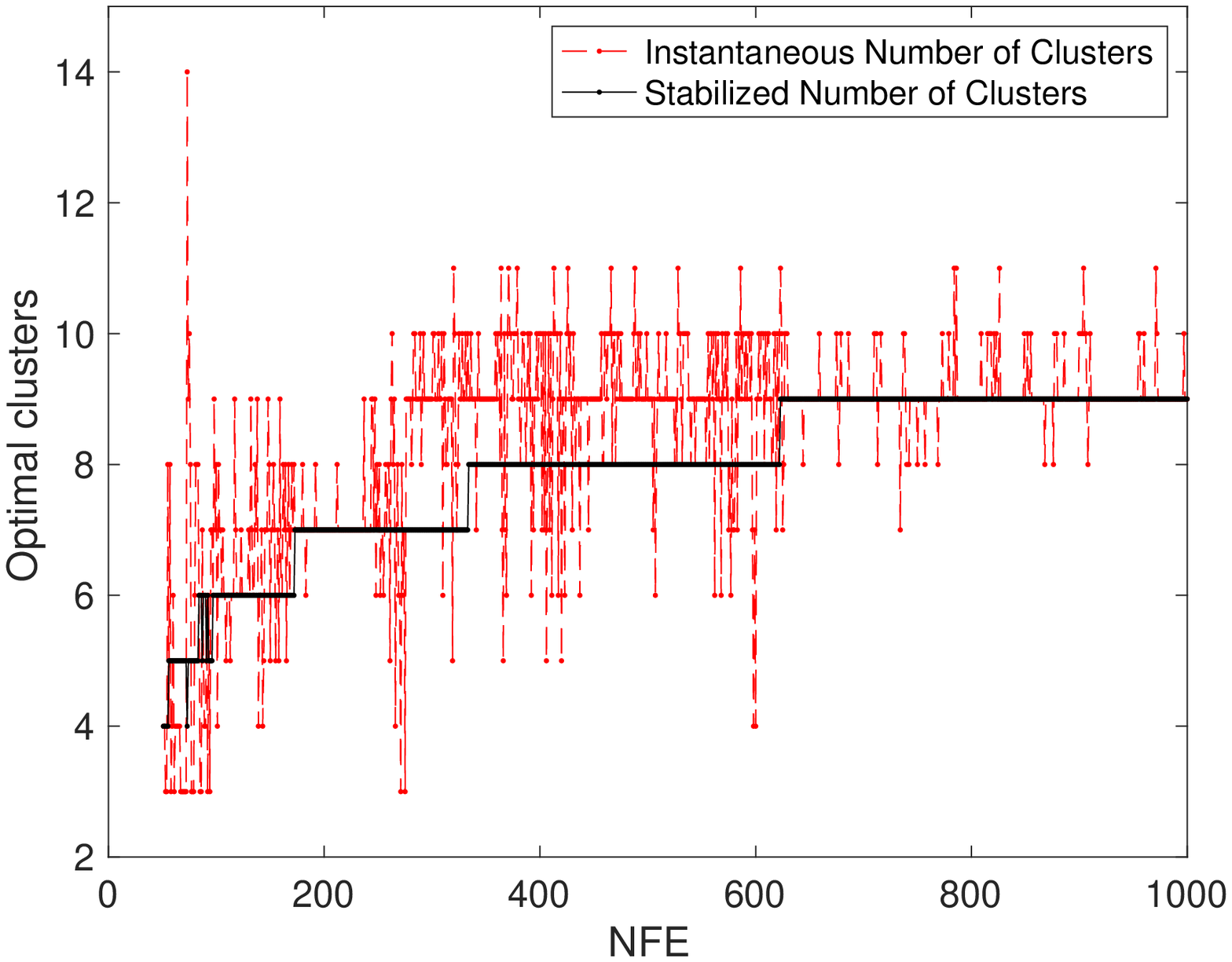}} 
\caption{(a) Samples explored by MOMO in objective space (b) Samples explored by MOMO in variable space and (c) Number of instantaneous and stabilized clusters $\{k^*, \bar{k}^*\}$ over the run}
\label{fig:Numfig}
\end{figure}

Let us illustrate the behavior of MOMO using the SYM-PART Simple problem~\cite{rudolph2007capabilities}. It is easy to visualize both the objective and the variable spaces for this problem as it involves two variables and two objectives. The PF is convex and there are $9$ equivalent PSS. We present the state of the parent population at $25\%$, $50\%$, $75\%$ and at the end of the computational budget for a particular run in Figure~\ref{fig:NumFig-XF}. In the left column of figures, each cluster in the variable space is shown in a different color. The same color scheme is also used to show its corresponding objective values in the right column. One can note that MOMO identified solutions in $8$ PSS locations out of $9$ within $25\%$ of $NFE_{max}$ as evident from Figure~\ref{fig:Numfig-X-25}. Beyond $50\%$ of $NFE_{max}$ onwards, it located solutions in all $9$ PSS although a closer look would reveal that at $50\%$ of $NFE_{max}$, it was working with a $\bar{k}^*=8$ as evident from Figure~\ref{fig:Numfig-X-50}~(8 clusters indicated in different colors). However, shortly after~(at approximately 600 evaluations), the stabilized number of clusters~$k$ increased to $9$ and maintained that value till the end of the run~(Figure~\ref{fig:Numfig-X-75} and Figure~\ref{fig:Numfig-X-100}). We also present all the solutions explored during the course of the run in Figure~\ref{fig:Numfig-1} and Figure~\ref{fig:Numfig-2}. One can notice higher intensity of sampling in and around $9$ locations that correspond to the PSS locations. The instantaneous and stabilized number of optimal clusters recorded throughout the process of evolution for this run is presented in Figure~\ref{fig:Numfig-3}. Again, it can be seen that although the instantaneous number of clusters show fluctuations, the stabilized value settles to the correct value of $9$ around $NFE = 600$.

\section{Numerical Experiments}
\label{sec:NR}

In this section, we present the experiments conducted on a wide range of problems, and the performance comparisons with six state-of-the-art peer-algorithms designed to solve MMOPs.  

\subsection{Test problems}
The proposed algorithm MOMO is tested on $21$ multi-modal multi-objective test problems~(MMF, SYM-PART simple,  SYM-PART rotated, Omni-test problem, MMMOP and IDMP). The number of equivalent PSS, number of variables~($D$) and number of objectives~($M$) for the problems are listed in Table~\ref{tab:PS table}. 

\begin{table}[]
\centering
\caption{Problems used in this study~(PSS = Number of equivalent Pareto subsets; D = Number of variables; M = Number of objectives)}
\label{tab:PS table}
\begin{tabular}{lcccl}
\hline
\textbf{Problem Name[Reference]} & \textbf{PSS} & \textbf{D} & \textbf{M} & \textbf{PF Geometry} \\ \hline
MMF1~\cite{yue2017multiobjective}                 & 2      & 2   & 2  & Convex     \\ 
MMF2~\cite{yue2017multiobjective}                 & 2   & 2   & 2  & Convex        \\ 
MMF3~\cite{yue2017multiobjective}                & 2    & 2   & 2  & Convex       \\ 
MMF4~\cite{yue2017multiobjective}                 & 4    & 2   & 2  & Concave      \\ 
MMF5~\cite{yue2017multiobjective}                 & 4     & 2   & 2  & Convex         \\ 
MMF6~\cite{yue2017multiobjective}                 & 4      & 2   & 2  & Convex        \\ 
MMF7~\cite{yue2017multiobjective}                 & 2      & 2   & 2  & Convex        \\ 
MMF8~\cite{yue2017multiobjective}                 & 4 & 2   & 2  & Concave          \\ 
SYM-PART Simple~\cite{rudolph2007capabilities}       & 9   & 2   & 2  & Convex        \\ 
SYM-PART Rotated~\cite{rudolph2007capabilities}      & 9   & 2   & 2  &  Convex         \\ 
Omni-test~\cite{deb2005omni}               & 9     & 2   & 2  & Convex      \\

MMMOP1A~\cite{liu2018multimodal}                 & 5      & 3   & 2  & Linear     \\ 
MMMOP2A~\cite{liu2018multimodal}                 & 6      & 3   & 2  & Concave     \\ 
MMMOP3A~\cite{liu2018multimodal}                 & 3      & 2   & 2  & Concave     \\ 
MMMOP4A~\cite{liu2018multimodal}                 & 4      & 2   & 2  & Concave     \\ 
MMMOP5A~\cite{liu2018multimodal}                 & 4      & 2   & 2  & Concave     \\ 
MMMOP6A~\cite{liu2018multimodal}                 & 2      & 2   & 2  & Concave     \\ 
IDMP-M2-T1~\cite{liu2019handling}                 & 2      & 2   & 2  & Linear     \\ 
IDMP-M2-T2~\cite{liu2019handling}                 & 2      & 2   & 2  & Linear     \\ 
IDMP-M2-T3~\cite{liu2019handling}                 & 2      & 2   & 2  & Linear     \\ 
IDMP-M2-T4~\cite{liu2019handling}                 & 2      & 2   & 2  & Linear     \\ \hline
\end{tabular}
\end{table}

\subsection{Compared algorithms and implementations}

The performance of the proposed algorithm is compared with algorithms that are available within the PlatEMO framework~(Version 3.3, released 2021/08/14)~\cite{tian2017platemo} and listed with the capability to deal with MMOPs. i.e. MO\_Ring\_PSO\_SCD, DN-NSGAII and TriMOEA-TA\&R and three other recent algorithms~(CPDEA, MMOEA/DC and MMEA-WI) for which the codes were made available by the respective authors. MOMO is implemented in MATLAB 2021b and all numerical experiments were performed on MacBook Pro(2019) with 2.3 GHz 8-Core Intel Core i9 processor and 32GB RAM. Codes for initialization, simulated binary crossover, polynomial mutation and non-dominated sorting used in MOMO were inherited from the PlatEMO framework~(2021) while $k-$means clustering and Silhouette index computation was based on in-built MATLAB 2021b toolboxes.

\subsection{Experimental settings}
We have used a population size of $50$ and allowed a computational budget of $1000$ function evaluations. Probability of SBX crossover is set to $1$ and the probability of polynomial mutation is set to $1/D$ for all algorithms with the distribution indices set to $20$ for both crossover and mutation. Statistical results are based on $31$ independent runs. The performance is judged using IGDX\cite{zhou2009approximating}, PSP\cite{yue2017multiobjective} and IGD\cite{wang2020adaptive} metrics that are widely used for assessing performance of multi-modal multi-objective optimization algorithms. For detailed definitions of the metrics, the readers are directed to the corresponding references. The implementations of IGD and IGDX metric computations are based on PlatEMO~(2021), while the codes of PSP were obtained from \cite{yue2017multiobjective}. 

Note that the current practice of benchmarking in MMOPs is based on metrics computed using the archive $\mathcal{A}$ of all evaluated solutions. We have used the same to maintain consistency in the comparisons with the peer-algorithms in this study. However, if one is interested in computing such metrics based on a final population or $N$ solutions, the population $P$ obtained at the end from Algo.~\ref{algo:base} can be considered. More involved schemes can also be applied. For example, if there are more than $N$ non-dominated solutions in $\mathcal{A}$, one can use the environmental selection mechanism presented in Section~\ref{subsec:env} to recursively delete the last solution until exactly $N$ solutions are left in the set. 

For SYM-PART Simple, SYM-PART Rotated, Omni-test problem and MMMOP3A, we have used a reference set containing $999$ solutions in PF and PS. For MMMOP2A, we used $1002$ solutions, while for all other problems we used $1000$ points in the reference sets for PF and PS. The reference PF and PS data and the source code of the algorithm is available for download from \url{http://www.mdolab.net/research_resources.html}. The size of the PS and PF is the same and both have been independently generated maintaining uniformity. These numbers are closest to $1000$ that conform to the characteristics, e.g. $9$ Pareto subsets with each subset containing $111$ points would yield a PS of $999$ solutions and for PF, $999$ solutions were generated uniformly. 

\subsection{Results}

\begin{table*}[!htb]
\centering
\caption{Mean and std. dev. across 31 runs based on IGDX. The best mean results are highlighted in grey. W/T/L represent the number of instances the compared algorithm was statistically better/equivalent/worse than MOMO using Wilcoxon Rank Sum (WRS) test.}
\label{tab:Performance comparison(IGDX)}
\tabcolsep 0.3mm
\renewcommand{\arraystretch}{1.3}
\begin{center}
\resizebox{\textwidth}{!}{
\begin{tabular}{cccccccc}
\hline
 & \multicolumn{1}{c}{\textbf{MOMO}} & \multicolumn{1}{c}{\textbf{MO\_Ring\_PSO\_SCD}} & \multicolumn{1}{c}{\textbf{DN-NSGAII}} & \multicolumn{1}{c}{\textbf{TriMOEA-TA\&R}} & \multicolumn{1}{c}{\textbf{CPDEA}} & \multicolumn{1}{c}{\textbf{MMOEA/DC}} & \multicolumn{1}{c}{\textbf{MMEA-WI}}\\\hline

\multirow{1}{*}{\textbf{MMF1}} & 4.59e-02 (9.21e-03) & \cellcolor{gray!25}4.31e-02 (4.54e-03) $\approx$ & 9.45e-02 (1.53e-02) $\downarrow$ & 9.48e-02 (1.26e-02) $\downarrow$ & 1.05e-01 (1.46e-02) $\downarrow$ & 6.93e-02 (9.71e-03) $\downarrow$ & 1.24e-01 (2.19e-02) $\downarrow$ \\
\hline

\multirow{1}{*}{\textbf{MMF2}} & \cellcolor{gray!25}3.48e-02 (6.82e-03) & 4.25e-02 (8.25e-03) $\downarrow$ & 6.90e-02 (9.89e-03) $\downarrow$ & 6.76e-02 (8.52e-03) $\downarrow$ & 1.02e-01 (2.29e-02) $\downarrow$ & 5.25e-02 (1.07e-02) $\downarrow$ & 1.80e-01 (6.13e-02) $\downarrow$ \\
\hline

\multirow{1}{*}{\textbf{MMF3}} & \cellcolor{gray!25}3.07e-02 (5.10e-03) & 3.84e-02 (8.65e-03) $\downarrow$ & 6.12e-02 (8.45e-03) $\downarrow$ & 6.05e-02 (8.96e-03) $\downarrow$ & 9.86e-02 (1.94e-02) $\downarrow$ & 4.57e-02 (7.66e-03) $\downarrow$ & 1.29e-01 (3.91e-02) $\downarrow$ \\
\hline

\multirow{1}{*}{\textbf{MMF4}} & \cellcolor{gray!25}3.95e-02 (4.40e-03) & 4.62e-02 (5.82e-03) $\downarrow$ & 6.09e-02 (1.06e-02) $\downarrow$ & 7.51e-02 (1.54e-02) $\downarrow$ & 5.07e-02 (5.23e-03) $\downarrow$ & 4.88e-02 (3.51e-03) $\downarrow$ & 5.87e-02 (9.49e-03) $\downarrow$ \\
\hline

\multirow{1}{*}{\textbf{MMF5}} & 5.54e-01 (1.15e-02) & \cellcolor{gray!25}5.45e-01 (5.79e-03) $\uparrow$ & 6.48e-01 (2.87e-02) $\downarrow$ & 6.53e-01 (2.68e-02) $\downarrow$ & 6.54e-01 (2.19e-02) $\downarrow$ & 6.11e-01 (2.38e-02) $\downarrow$ & 7.06e-01 (4.26e-02) $\downarrow$ \\
\hline

\multirow{1}{*}{\textbf{MMF6}} & 5.48e-01 (2.47e-02) & \cellcolor{gray!25}5.37e-01 (4.52e-03) $\uparrow$ & 6.19e-01 (3.12e-02) $\downarrow$ & 6.25e-01 (3.23e-02) $\downarrow$ & 6.23e-01 (1.75e-02) $\downarrow$ & 5.93e-01 (2.14e-02) $\downarrow$ & 6.82e-01 (6.16e-02) $\downarrow$ \\
\hline

\multirow{1}{*}{\textbf{MMF7}} & \cellcolor{gray!25}3.42e-02 (4.33e-03) & 3.65e-02 (3.10e-03) $\downarrow$ & 4.56e-02 (7.04e-03) $\downarrow$ & 6.12e-02 (1.03e-02) $\downarrow$ & 4.91e-02 (4.16e-03) $\downarrow$ & 4.34e-02 (4.74e-03) $\downarrow$ & 5.17e-02 (9.26e-03) $\downarrow$ \\
\hline

\multirow{1}{*}{\textbf{MMF8}} & \cellcolor{gray!25}1.44e-01 (2.55e-02) & 1.49e-01 (2.04e-02) $\approx$ & 2.42e-01 (5.72e-02) $\downarrow$ & 2.62e-01 (5.30e-02) $\downarrow$ & 2.30e-01 (5.83e-02) $\downarrow$ & 1.60e-01 (2.92e-02) $\downarrow$ & 2.75e-01 (8.40e-02) $\downarrow$ \\
\hline

\multirow{1}{*}{\textbf{SYM-PART1}} & \cellcolor{gray!25}1.47e-01 (1.07e-01) & 5.40e-01 (1.00e-01) $\downarrow$ & 6.59e-01 (3.63e-01) $\downarrow$ & 3.01e-01 (2.58e-01) $\downarrow$ & 2.89e+00 (1.31e+00) $\downarrow$ & 2.45e-01 (1.22e-01) $\downarrow$ & 1.08e+00 (1.05e+00) $\downarrow$ \\
\hline

\multirow{1}{*}{\textbf{SYM-PART2}} & \cellcolor{gray!25}4.11e-01 (2.20e-01) & 6.11e-01 (1.01e-01) $\downarrow$ & 1.11e+00 (3.12e-01) $\downarrow$ & 1.23e+00 (3.80e-01) $\downarrow$ & 3.83e+00 (1.69e+00) $\downarrow$ & 6.05e-01 (2.64e-01) $\downarrow$ & 2.68e+00 (1.21e+00) $\downarrow$ \\
\hline

\multirow{1}{*}{\textbf{Omni-test}} & \cellcolor{gray!25}5.65e-02 (7.19e-03) & 9.46e-02 (1.09e-02) $\downarrow$ & 1.21e-01 (7.03e-02) $\downarrow$ & 1.35e-01 (5.28e-02) $\downarrow$ & 1.33e-01 (7.16e-02) $\downarrow$ & 6.64e-02 (1.03e-02) $\downarrow$ & 9.33e-02 (6.66e-02) $\downarrow$ \\
\hline

\multirow{1}{*}{\textbf{MMMOP1A}} & 8.16e-02 (1.96e-02) & \cellcolor{gray!25}6.73e-02 (6.49e-03) $\uparrow$ & 1.06e-01 (2.02e-02) $\downarrow$ & 9.98e-02 (1.90e-02) $\downarrow$ & 2.05e-01 (4.48e-02) $\downarrow$ & 1.09e-01 (1.56e-02) $\downarrow$ & 2.42e-01 (3.92e-02) $\downarrow$ \\
\hline

\multirow{1}{*}{\textbf{MMMOP2A}} & \cellcolor{gray!25}5.72e-02 (5.25e-02) & 1.33e-01 (4.18e-02) $\downarrow$ & 1.29e-01 (6.34e-02) $\downarrow$ & 7.76e-02 (6.62e-02) $\approx$ & 9.63e-02 (9.59e-02) $\downarrow$ & 7.53e-02 (6.62e-02) $\approx$ & 2.68e-01 (2.49e-01) $\downarrow$ \\
\hline

\multirow{1}{*}{\textbf{MMMOP3A}} & 1.29e-02 (1.95e-03) & 2.19e-02 (3.64e-03) $\downarrow$ & \cellcolor{gray!25}9.63e-03 (8.51e-04) $\uparrow$ & 1.13e-02 (2.07e-03) $\uparrow$ & 1.51e-02 (1.82e-03) $\downarrow$ & 1.61e-02 (2.23e-03) $\downarrow$ & 1.18e-02 (1.57e-03) $\uparrow$ \\
\hline

\multirow{1}{*}{\textbf{MMMOP4A}} & \cellcolor{gray!25}1.49e-02 (8.39e-03) & 1.49e-02 (2.12e-03) $\approx$ & 1.57e-02 (1.20e-02) $\approx$ & 2.08e-02 (1.47e-02) $\downarrow$ & 2.97e-02 (2.08e-02) $\downarrow$ & 1.99e-02 (1.20e-02) $\downarrow$ & 6.13e-02 (3.59e-02) $\downarrow$ \\
\hline

\multirow{1}{*}{\textbf{MMMOP5A}} & \cellcolor{gray!25}1.54e-02 (9.11e-03) & 1.58e-02 (2.27e-03) $\downarrow$ & 2.26e-02 (1.71e-02) $\downarrow$ & 2.03e-02 (9.99e-03) $\downarrow$ & 3.40e-02 (2.32e-02) $\downarrow$ & 2.29e-02 (1.23e-02) $\downarrow$ & 6.24e-02 (2.90e-02) $\downarrow$ \\
\hline

\multirow{1}{*}{\textbf{MMMOP6A}} & 2.17e-02 (5.64e-03) & \cellcolor{gray!25}1.84e-02 (1.78e-03) $\uparrow$ & 3.07e-02 (5.92e-03) $\downarrow$ & 3.43e-02 (7.15e-03) $\downarrow$ & 2.42e-02 (5.24e-03) $\downarrow$ & 2.02e-02 (4.40e-03) $\approx$ & 3.11e-02 (9.31e-03) $\downarrow$ \\
\hline

\multirow{1}{*}{\textbf{IDMPM2T1}} & 7.29e-03 (2.98e-03) & 2.64e-02 (1.95e-02) $\downarrow$ & 2.92e-02 (2.95e-02) $\downarrow$ & 3.14e-02 (2.84e-02) $\downarrow$ & 7.68e-02 (1.97e-01) $\downarrow$ & 1.32e-02 (1.65e-02) $\approx$ & \cellcolor{gray!25}6.61e-03 (5.27e-03) $\uparrow$ \\
\hline

\multirow{1}{*}{\textbf{IDMPM2T2}} & \cellcolor{gray!25}5.51e-03 (1.21e-03) & 1.96e-02 (8.77e-03) $\downarrow$ & 2.71e-02 (2.71e-02) $\downarrow$ & 3.82e-02 (3.42e-02) $\downarrow$ & 3.93e-02 (1.22e-01) $\downarrow$ & 1.07e-02 (9.06e-03) $\downarrow$ & 4.05e-02 (1.18e-01) $\downarrow$ \\
\hline

\multirow{1}{*}{\textbf{IDMPM2T3}} & \cellcolor{gray!25}1.97e-02 (8.82e-03) & 3.40e-02 (1.84e-02) $\downarrow$ & 4.71e-02 (3.24e-02) $\downarrow$ & 5.28e-02 (3.46e-02) $\downarrow$ & 1.32e-01 (1.80e-01) $\downarrow$ & 4.52e-02 (2.80e-02) $\downarrow$ & 1.51e-01 (1.95e-01) $\downarrow$ \\
\hline

\multirow{1}{*}{\textbf{IDMPM2T4}} & \cellcolor{gray!25}1.33e-02 (1.02e-02) & 2.76e-02 (1.03e-02) $\downarrow$ & 3.92e-02 (3.38e-02) $\downarrow$ & 3.95e-02 (2.95e-02) $\downarrow$ & 3.25e-01 (3.01e-01) $\downarrow$ & 3.01e-02 (2.58e-02) $\downarrow$ & 3.15e-01 (2.56e-01) $\downarrow$ \\
\hline

\multirow{1}{*}{\textbf{W/T/L}} &  & 4/3/14 & 1/1/19 & 1/1/19 & 0/0/21 & 0/3/18 & 2/0/19 \\\hline

\end{tabular}}
\end{center}
\end{table*}

The performance of the algorithm based on IGDX is presented in Table~\ref{tab:Performance comparison(IGDX)} with the mean and the standard deviation reported for all algorithms computed based on $31$ independent runs. Wilcoxon rank-sum results of significance test is indicated using the symbols $\uparrow, \approx, \downarrow$, where the compared algorithm performs significantly better, equivalent, or worse than MOMO. In the last row, the cumulative Win/Tie/Loss~(W/T/L) numbers indicates that the number of times the compared algorithm's performance is better, equivalent, or worse than the proposed algorithm. From the results, it can be seen that out of $21$ problems, MO\_Ring\_PSO\_SCD, DN-NSGAII, TriMOEA-TA\&R, CPDEA, MMOEA/DC and MMEA-WI record $14$, $19$, $19$, $21$, $18$ and $19$ losses, respectively, against MOMO. The best mean results are marked in grey. Overall, MOMO significantly outperforms the peers in achieving the best IGDX results across all algorithms, as well as in the pairwise comparisons. 

PSP is yet another metric that is commonly used to assess the performance of such algorithms. The mean and standard deviation of PSP metric across 31 runs are presented in Table~\ref{tab:Performance comparison(PSP)} for all algorithms. Once again the results of significance test based on PSP indicates that MO\_Ring\_PSO\_SCD, DN-NSGAII, TriMOEA-TA\&R, CPDEA, MMOEA/DC and MMEA-WI record $15$, $19$, $20$, $21$, $18$ and $19$ losses against MOMO.

While both the above measures assess the performance based on variable space~(obtained PS against the reference PS), the results of IGD presented in Table~\ref{tab:Performance comparison(IGD)} reflects on its ability to deliver converged and well distributed set of solutions on or close to the PF in the objective space.  The results of significance test based on IGD indicates that MO\_Ring\_PSO\_SCD, DN-NSGAII, TriMOEA-TA\&R, CPDEA, MMOEA/DC and MMEA-WI record $18$, $6$, $12$, $7$, $9$ and $12$ losses against MOMO. MOMO records more wins than losses over all algorithms except for DN-NSGAII which has the best result based on IGD~(notably, with significantly inferior results in IGDX, PSP metrics). 

\begin{table*}[!htb]
\centering
\caption{Mean and std. dev. across 31 runs based on PSP. The best mean results are highlighted in grey. W/T/L represent the number of instances the compared algorithm was statistically better/equivalent/worse than MOMO using Wilcoxon Rank Sum (WRS) test.}
\label{tab:Performance comparison(PSP)}
\tabcolsep 0.3mm
\renewcommand{\arraystretch}{1.3}
\begin{center}
\resizebox{\textwidth}{!}{
\begin{tabular}{cccccccc}
\hline
 & \multicolumn{1}{c}{\textbf{MOMO}} & \multicolumn{1}{c}{\textbf{MO\_Ring\_PSO\_SCD}} & \multicolumn{1}{c}{\textbf{DN-NSGAII}} & \multicolumn{1}{c}{\textbf{TriMOEA-TA\&R}} & \multicolumn{1}{c}{\textbf{CPDEA}} & \multicolumn{1}{c}{\textbf{MMOEA/DC}} & \multicolumn{1}{c}{\textbf{MMEA-WI}}\\\hline

\multirow{1}{*}{\textbf{MMF1}} & 2.17e+01 (4.23e+00) & \cellcolor{gray!25}2.24e+01 (2.51e+00) $\approx$ & 9.94e+00 (2.05e+00) $\downarrow$ & 1.00e+01 (1.70e+00) $\downarrow$ & 9.49e+00 (1.35e+00) $\downarrow$ & 1.43e+01 (2.02e+00) $\downarrow$ & 8.00e+00 (1.47e+00) $\downarrow$ \\
\hline

\multirow{1}{*}{\textbf{MMF2}} & \cellcolor{gray!25}2.59e+01 (6.44e+00) & 1.89e+01 (5.11e+00) $\downarrow$ & 1.08e+01 (3.04e+00) $\downarrow$ & 1.10e+01 (2.38e+00) $\downarrow$ & 9.63e+00 (2.23e+00) $\downarrow$ & 1.69e+01 (4.38e+00) $\downarrow$ & 5.09e+00 (2.10e+00) $\downarrow$ \\
\hline

\multirow{1}{*}{\textbf{MMF3}} & \cellcolor{gray!25}2.79e+01 (5.81e+00) & 2.09e+01 (6.90e+00) $\downarrow$ & 1.20e+01 (2.79e+00) $\downarrow$ & 1.20e+01 (3.62e+00) $\downarrow$ & 1.00e+01 (2.09e+00) $\downarrow$ & 1.93e+01 (4.93e+00) $\downarrow$ & 6.87e+00 (2.73e+00) $\downarrow$ \\
\hline

\multirow{1}{*}{\textbf{MMF4}} & \cellcolor{gray!25}2.52e+01 (2.82e+00) & 2.13e+01 (2.69e+00) $\downarrow$ & 1.64e+01 (3.02e+00) $\downarrow$ & 1.34e+01 (2.93e+00) $\downarrow$ & 1.98e+01 (1.89e+00) $\downarrow$ & 2.04e+01 (1.51e+00) $\downarrow$ & 1.73e+01 (2.59e+00) $\downarrow$ \\
\hline

\multirow{1}{*}{\textbf{MMF5}} & 1.23e+00 (5.34e-02) & \cellcolor{gray!25}1.24e+00 (3.66e-02) $\approx$ & 9.96e-01 (9.11e-02) $\downarrow$ & 9.57e-01 (1.07e-01) $\downarrow$ & 1.07e+00 (4.18e-02) $\downarrow$ & 1.13e+00 (5.61e-02) $\downarrow$ & 9.70e-01 (7.98e-02) $\downarrow$ \\
\hline

\multirow{1}{*}{\textbf{MMF6}} & 1.24e+00 (9.27e-02) & \cellcolor{gray!25}1.26e+00 (3.15e-02) $\approx$ & 1.06e+00 (8.42e-02) $\downarrow$ & 1.04e+00 (8.84e-02) $\downarrow$ & 1.12e+00 (3.82e-02) $\downarrow$ & 1.15e+00 (6.20e-02) $\downarrow$ & 1.00e+00 (1.14e-01) $\downarrow$ \\
\hline

\multirow{1}{*}{\textbf{MMF7}} & \cellcolor{gray!25}2.78e+01 (4.11e+00) & 2.60e+01 (2.63e+00) $\downarrow$ & 1.92e+01 (3.88e+00) $\downarrow$ & 1.34e+01 (3.13e+00) $\downarrow$ & 1.96e+01 (2.19e+00) $\downarrow$ & 2.17e+01 (2.70e+00) $\downarrow$ & 1.81e+01 (3.78e+00) $\downarrow$ \\
\hline

\multirow{1}{*}{\textbf{MMF8}} & \cellcolor{gray!25}6.71e+00 (1.16e+00) & 6.39e+00 (9.43e-01) $\approx$ & 3.91e+00 (1.09e+00) $\downarrow$ & 3.61e+00 (7.81e-01) $\downarrow$ & 4.37e+00 (8.99e-01) $\downarrow$ & 6.00e+00 (1.24e+00) $\downarrow$ & 3.73e+00 (1.02e+00) $\downarrow$ \\
\hline

\multirow{1}{*}{\textbf{SYM-PART1}} & \cellcolor{gray!25}7.88e+00 (2.01e+00) & 1.81e+00 (4.25e-01) $\downarrow$ & 1.74e+00 (1.42e+00) $\downarrow$ & 5.14e+00 (3.80e+00) $\downarrow$ & 3.46e-01 (2.50e-01) $\downarrow$ & 4.56e+00 (1.26e+00) $\downarrow$ & 2.54e+00 (2.09e+00) $\downarrow$ \\
\hline

\multirow{1}{*}{\textbf{SYM-PART2}} & \cellcolor{gray!25}2.75e+00 (1.15e+00) & 1.37e+00 (3.64e-01) $\downarrow$ & 6.11e-01 (2.52e-01) $\downarrow$ & 5.70e-01 (3.44e-01) $\downarrow$ & 1.63e-01 (9.16e-02) $\downarrow$ & 1.69e+00 (7.88e-01) $\downarrow$ & 5.50e-01 (7.67e-01) $\downarrow$ \\
\hline

\multirow{1}{*}{\textbf{Omni-test}} & \cellcolor{gray!25}1.77e+01 (2.35e+00) & 1.05e+01 (1.31e+00) $\downarrow$ & 1.01e+01 (5.03e+00) $\downarrow$ & 7.79e+00 (3.33e+00) $\downarrow$ & 8.87e+00 (2.90e+00) $\downarrow$ & 1.52e+01 (1.94e+00) $\downarrow$ & 1.38e+01 (5.00e+00) $\downarrow$ \\
\hline

\multirow{1}{*}{\textbf{MMMOP1A}} & 1.12e+01 (4.08e+00) & \cellcolor{gray!25}1.37e+01 (1.88e+00) $\uparrow$ & 5.00e+00 (3.47e+00) $\downarrow$ & 2.68e-06 (1.07e-05) $\downarrow$ & 3.70e+00 (1.83e+00) $\downarrow$ & 8.49e+00 (1.80e+00) $\downarrow$ & 3.39e+00 (1.40e+00) $\downarrow$ \\
\hline

\multirow{1}{*}{\textbf{MMMOP2A}} & \cellcolor{gray!25}3.06e+01 (2.47e+01) & 5.81e+00 (3.12e+00) $\downarrow$ & 6.66e+00 (8.51e+00) $\downarrow$ & 2.71e-02 (1.20e-01) $\downarrow$ & 1.74e+01 (1.38e+01) $\downarrow$ & 2.31e+01 (1.82e+01) $\approx$ & 6.75e+00 (7.25e+00) $\downarrow$ \\
\hline

\multirow{1}{*}{\textbf{MMMOP3A}} & 7.89e+01 (1.29e+01) & 4.61e+01 (7.67e+00) $\downarrow$ & \cellcolor{gray!25}1.04e+02 (8.88e+00) $\uparrow$ & 9.06e+01 (1.59e+01) $\uparrow$ & 6.70e+01 (8.32e+00) $\downarrow$ & 6.33e+01 (8.88e+00) $\downarrow$ & 8.61e+01 (1.15e+01) $\uparrow$ \\
\hline

\multirow{1}{*}{\textbf{MMMOP4A}} & 8.55e+01 (5.22e+01) & 5.57e+01 (1.11e+01) $\downarrow$ & \cellcolor{gray!25}9.29e+01 (7.29e+01) $\approx$ & 6.16e+01 (3.78e+01) $\downarrow$ & 5.63e+01 (4.73e+01) $\downarrow$ & 6.14e+01 (3.24e+01) $\downarrow$ & 2.10e+01 (1.89e+01) $\downarrow$ \\
\hline

\multirow{1}{*}{\textbf{MMMOP5A}} & \cellcolor{gray!25}8.56e+01 (4.37e+01) & 5.62e+01 (1.06e+01) $\downarrow$ & 6.96e+01 (5.54e+01) $\downarrow$ & 5.86e+01 (2.90e+01) $\downarrow$ & 4.63e+01 (3.38e+01) $\downarrow$ & 5.48e+01 (2.51e+01) $\downarrow$ & 1.99e+01 (1.13e+01) $\downarrow$ \\
\hline

\multirow{1}{*}{\textbf{MMMOP6A}} & 4.80e+01 (1.13e+01) & \cellcolor{gray!25}5.38e+01 (5.43e+00) $\uparrow$ & 3.28e+01 (7.20e+00) $\downarrow$ & 2.95e+01 (7.10e+00) $\downarrow$ & 4.25e+01 (6.69e+00) $\downarrow$ & 5.09e+01 (9.18e+00) $\approx$ & 3.39e+01 (8.07e+00) $\downarrow$ \\
\hline

\multirow{1}{*}{\textbf{IDMPM2T1}} & 1.56e+02 (5.78e+01) & 3.02e+01 (3.37e+01) $\downarrow$ & 3.98e+01 (7.37e+01) $\downarrow$ & 5.15e+01 (1.03e+02) $\downarrow$ & 8.08e+01 (6.69e+01) $\downarrow$ & 1.25e+02 (7.40e+01) $\approx$ & \cellcolor{gray!25}1.81e+02 (7.32e+01) $\uparrow$ \\
\hline

\multirow{1}{*}{\textbf{IDMPM2T2}} & \cellcolor{gray!25}1.88e+02 (4.01e+01) & 3.82e+01 (2.93e+01) $\downarrow$ & 5.50e+01 (1.01e+02) $\downarrow$ & 3.15e+01 (7.34e+01) $\downarrow$ & 1.05e+02 (6.13e+01) $\downarrow$ & 1.12e+02 (6.76e+01) $\downarrow$ & 1.20e+02 (8.94e+01) $\downarrow$ \\
\hline

\multirow{1}{*}{\textbf{IDMPM2T3}} & \cellcolor{gray!25}5.27e+01 (2.04e+01) & 1.81e+01 (1.99e+01) $\downarrow$ & 1.66e+01 (2.08e+01) $\downarrow$ & 1.44e+01 (2.05e+01) $\downarrow$ & 1.26e+01 (1.30e+01) $\downarrow$ & 2.02e+01 (2.02e+01) $\downarrow$ & 1.47e+01 (1.39e+01) $\downarrow$ \\
\hline

\multirow{1}{*}{\textbf{IDMPM2T4}} & \cellcolor{gray!25}8.30e+01 (7.73e+01) & 6.99e+00 (1.97e+01) $\downarrow$ & 1.20e+01 (6.40e+01) $\downarrow$ & 1.44e+01 (5.42e+01) $\downarrow$ & 2.02e+01 (4.70e+01) $\downarrow$ & 4.20e+01 (6.88e+01) $\downarrow$ & 7.87e+00 (2.85e+01) $\downarrow$ \\
\hline

\multirow{1}{*}{\textbf{W/T/L}} &  & 2/4/15 & 1/1/19 & 1/0/20 & 0/0/21 & 0/3/18 & 2/0/19 \\\hline

\end{tabular}}
\end{center}
\end{table*}

\begin{table*}[!htb]
\centering
\caption{Mean and std. dev. across 31 runs based on IGD. The best mean results are highlighted in grey. W/T/L represent the number of instances the compared algorithm was statistically better/equivalent/worse than MOMO using Wilcoxon Rank-Sum test.}
\label{tab:Performance comparison(IGD)}
\tabcolsep 0.3mm
\renewcommand{\arraystretch}{1.3}
\begin{center}
\resizebox{\textwidth}{!}{
\begin{tabular}{cccccccc}
\hline
 & \multicolumn{1}{c}{\textbf{MOMO}} & \multicolumn{1}{c}{\textbf{MO\_Ring\_PSO\_SCD}} & \multicolumn{1}{c}{\textbf{DN-NSGAII}} & \multicolumn{1}{c}{\textbf{TriMOEA-TA\&R}} & \multicolumn{1}{c}{\textbf{CPDEA}} & \multicolumn{1}{c}{\textbf{MMOEA/DC}} & \multicolumn{1}{c}{\textbf{MMEA-WI}}\\\hline

\multirow{1}{*}{\textbf{MMF1}} & 1.23e-02 (4.70e-03) & 1.10e-02 (1.73e-03) $\approx$ & 1.68e-02 (9.25e-03) $\downarrow$ & 1.77e-02 (7.48e-03) $\downarrow$ & \cellcolor{gray!25}1.07e-02 (1.16e-03) $\approx$ & 1.11e-02 (2.12e-03) $\approx$ & 1.26e-02 (2.29e-03) $\downarrow$ \\
\hline

\multirow{1}{*}{\textbf{MMF2}} & 7.57e-02 (2.17e-02) & 1.30e-01 (4.74e-02) $\downarrow$ & 1.06e-01 (4.81e-02) $\downarrow$ & 9.80e-02 (3.16e-02) $\downarrow$ & \cellcolor{gray!25}7.25e-02 (2.08e-02) $\approx$ & 7.37e-02 (2.44e-02) $\approx$ & 1.23e-01 (5.88e-02) $\downarrow$ \\
\hline

\multirow{1}{*}{\textbf{MMF3}} & 6.81e-02 (2.38e-02) & 1.21e-01 (5.39e-02) $\downarrow$ & 9.54e-02 (3.54e-02) $\downarrow$ & 8.94e-02 (3.35e-02) $\downarrow$ & \cellcolor{gray!25}6.05e-02 (1.53e-02) $\approx$ & 6.71e-02 (1.90e-02) $\approx$ & 1.00e-01 (3.22e-02) $\downarrow$ \\
\hline

\multirow{1}{*}{\textbf{MMF4}} & \cellcolor{gray!25}6.02e-03 (8.83e-04) & 7.47e-03 (7.53e-04) $\downarrow$ & 6.38e-03 (1.08e-03) $\approx$ & 9.64e-03 (1.73e-03) $\downarrow$ & 7.48e-03 (6.42e-04) $\downarrow$ & 7.31e-03 (6.04e-04) $\downarrow$ & 6.86e-03 (7.82e-04) $\downarrow$ \\
\hline

\multirow{1}{*}{\textbf{MMF5}} & 1.20e-02 (2.79e-03) & \cellcolor{gray!25}1.01e-02 (1.22e-03) $\uparrow$ & 1.31e-02 (4.45e-03) $\approx$ & 1.71e-02 (4.40e-03) $\downarrow$ & 1.10e-02 (1.48e-03) $\approx$ & 1.10e-02 (1.44e-03) $\approx$ & 1.34e-02 (2.48e-03) $\downarrow$ \\
\hline

\multirow{1}{*}{\textbf{MMF6}} & 9.34e-03 (1.25e-03) & \cellcolor{gray!25}8.59e-03 (8.46e-04) $\uparrow$ & 9.38e-03 (1.98e-03) $\approx$ & 1.29e-02 (3.65e-03) $\downarrow$ & 9.38e-03 (1.33e-03) $\approx$ & 9.34e-03 (1.45e-03) $\approx$ & 1.05e-02 (3.18e-03) $\approx$ \\
\hline

\multirow{1}{*}{\textbf{MMF7}} & 6.20e-03 (6.92e-04) & 6.90e-03 (6.45e-04) $\downarrow$ & \cellcolor{gray!25}5.96e-03 (6.73e-04) $\uparrow$ & 9.70e-03 (2.41e-03) $\downarrow$ & 7.35e-03 (5.87e-04) $\downarrow$ & 6.66e-03 (6.25e-04) $\downarrow$ & 6.65e-03 (6.86e-04) $\downarrow$ \\
\hline

\multirow{1}{*}{\textbf{MMF8}} & \cellcolor{gray!25}7.35e-02 (1.70e-03) & 7.81e-02 (1.82e-03) $\downarrow$ & 7.38e-02 (1.04e-03) $\downarrow$ & 7.38e-02 (1.82e-03) $\approx$ & 7.59e-02 (1.44e-03) $\downarrow$ & 7.43e-02 (9.99e-04) $\downarrow$ & 7.44e-02 (1.07e-03) $\downarrow$ \\
\hline

\multirow{1}{*}{\textbf{SYM-PART1}} & \cellcolor{gray!25}7.83e-03 (1.22e-03) & 4.56e-02 (1.01e-02) $\downarrow$ & 8.75e-03 (1.03e-03) $\downarrow$ & 1.21e-02 (1.89e-03) $\downarrow$ & 2.76e-02 (5.84e-03) $\downarrow$ & 1.42e-02 (2.21e-03) $\downarrow$ & 1.39e-02 (2.56e-03) $\downarrow$ \\
\hline

\multirow{1}{*}{\textbf{SYM-PART2}} & \cellcolor{gray!25}1.32e-02 (2.25e-03) & 5.00e-02 (1.10e-02) $\downarrow$ & 1.38e-02 (2.23e-03) $\approx$ & 1.72e-02 (3.63e-03) $\downarrow$ & 3.20e-02 (6.40e-03) $\downarrow$ & 1.87e-02 (3.65e-03) $\downarrow$ & 1.81e-02 (2.91e-03) $\downarrow$ \\
\hline

\multirow{1}{*}{\textbf{Omni-test}} & \cellcolor{gray!25}6.74e-03 (9.29e-04) & 1.81e-02 (2.06e-03) $\downarrow$ & 7.10e-03 (6.86e-04) $\downarrow$ & 9.93e-03 (1.86e-03) $\downarrow$ & 1.38e-02 (1.95e-03) $\downarrow$ & 1.14e-02 (1.23e-03) $\downarrow$ & 7.99e-03 (9.09e-04) $\downarrow$ \\
\hline

\multirow{1}{*}{\textbf{MMMOP1A}} & 6.62e-01 (3.77e-01) & 2.96e+00 (2.42e+00) $\downarrow$ & \cellcolor{gray!25}4.94e-01 (5.59e-01) $\uparrow$ & 5.18e-01 (3.81e-01) $\uparrow$ & 6.65e-01 (6.21e-01) $\approx$ & 1.03e+00 (6.32e-01) $\downarrow$ & 1.41e+00 (1.09e+00) $\downarrow$ \\
\hline

\multirow{1}{*}{\textbf{MMMOP2A}} & 1.90e-01 (2.44e-01) & 6.38e-01 (1.93e-01) $\downarrow$ & 4.10e-01 (3.76e-01) $\approx$ & 3.42e-01 (3.35e-01) $\downarrow$ & \cellcolor{gray!25}8.42e-02 (6.32e-03) $\approx$ & 2.55e-01 (3.15e-01) $\approx$ & 4.50e-01 (3.79e-01) $\downarrow$ \\
\hline

\multirow{1}{*}{\textbf{MMMOP3A}} & 7.13e-02 (2.26e-04) & 7.30e-02 (6.01e-04) $\downarrow$ & \cellcolor{gray!25}7.10e-02 (2.05e-04) $\uparrow$ & 7.13e-02 (2.29e-04) $\approx$ & 7.19e-02 (2.23e-04) $\downarrow$ & 7.18e-02 (3.41e-04) $\downarrow$ & 7.13e-02 (1.88e-04) $\approx$ \\
\hline

\multirow{1}{*}{\textbf{MMMOP4A}} & 2.48e-01 (2.80e-01) & 3.64e-01 (1.31e-01) $\downarrow$ & \cellcolor{gray!25}1.09e-01 (1.07e-01) $\uparrow$ & 1.58e-01 (2.67e-01) $\uparrow$ & 1.20e-01 (5.66e-02) $\approx$ & 2.20e-01 (2.46e-01) $\approx$ & 2.46e-01 (2.35e-01) $\approx$ \\
\hline

\multirow{1}{*}{\textbf{MMMOP5A}} & 1.64e-01 (1.27e-01) & 3.79e-01 (1.90e-01) $\downarrow$ & 2.13e-01 (2.92e-01) $\uparrow$ & \cellcolor{gray!25}9.53e-02 (2.64e-02) $\uparrow$ & 1.04e-01 (3.00e-02) $\uparrow$ & 2.38e-01 (2.74e-01) $\approx$ & 2.30e-01 (2.21e-01) $\approx$ \\
\hline

\multirow{1}{*}{\textbf{MMMOP6A}} & 7.53e-02 (3.70e-03) & 7.52e-02 (1.40e-03) $\downarrow$ & 7.48e-02 (1.93e-03) $\approx$ & 7.70e-02 (3.43e-03) $\downarrow$ & 7.38e-02 (7.84e-04) $\approx$ & \cellcolor{gray!25}7.36e-02 (1.12e-03) $\uparrow$ & 7.37e-02 (1.01e-03) $\approx$ \\
\hline

\multirow{1}{*}{\textbf{IDMPM2T1}} & 3.91e-02 (1.59e-02) & 7.60e-02 (1.66e-02) $\downarrow$ & 1.47e-02 (4.52e-03) $\uparrow$ & \cellcolor{gray!25}1.34e-02 (5.62e-03) $\uparrow$ & 2.83e-02 (8.02e-03) $\uparrow$ & 3.13e-02 (8.23e-03) $\uparrow$ & 2.47e-02 (6.62e-03) $\uparrow$ \\
\hline

\multirow{1}{*}{\textbf{IDMPM2T2}} & 2.27e-02 (9.32e-03) & 9.82e-02 (4.34e-02) $\downarrow$ & 1.10e-02 (4.46e-03) $\uparrow$ & \cellcolor{gray!25}1.01e-02 (2.94e-03) $\uparrow$ & 2.39e-02 (8.10e-03) $\approx$ & 2.64e-02 (1.04e-02) $\downarrow$ & 2.16e-02 (7.28e-03) $\approx$ \\
\hline

\multirow{1}{*}{\textbf{IDMPM2T3}} & 5.30e-02 (1.82e-02) & 8.07e-02 (3.17e-02) $\downarrow$ & 1.41e-02 (1.64e-02) $\uparrow$ & \cellcolor{gray!25}1.35e-02 (1.61e-02) $\uparrow$ & 2.92e-02 (9.52e-03) $\uparrow$ & 2.50e-02 (9.85e-03) $\uparrow$ & 5.64e-02 (5.38e-02) $\uparrow$ \\
\hline

\multirow{1}{*}{\textbf{IDMPM2T4}} & 8.65e-02 (4.67e-02) & 3.44e-01 (2.26e-01) $\downarrow$ & \cellcolor{gray!25}1.50e-02 (7.99e-03) $\uparrow$ & 2.08e-02 (1.63e-02) $\uparrow$ & 8.06e-02 (9.25e-02) $\uparrow$ & 5.06e-02 (2.72e-02) $\uparrow$ & 4.90e-02 (4.18e-02) $\uparrow$ \\
\hline

\multirow{1}{*}{\textbf{W/T/L}} &  & 2/1/18 & 9/6/6 & 7/2/12 & 4/10/7 & 4/8/9 & 3/6/12 \\\hline

\end{tabular}}
\end{center}
\end{table*}

The above experiments, conducted across a wide range of problems, and supported by statistical significance tests using multiple established metrics, clearly reflect on the ability of MOMO to deliver competitive results. Notably, MOMO operates on a very simple design compared to other peer-algorithms, with no additional parameters that require tuning. We hope that this study would encourage further development of simpler, algorithms to solve MMOPs, which could further encourage their uptake by the practitioners. 

\section{Conclusions}
\label{sec:SC}

In this paper we have introduced a simple and an efficient algorithm for multi-modal multi-objective optimization. The algorithm is based on a steady-state framework and aims to solve MMOPs with a limited computing budget of $1000$ function evaluations. The population is clustered in the variable space into clusters based on the running average of the optimum number of clusters~($\bar{k}^*$) determined using Silhouette index. Two parents selected from the smallest clusters undergo recombination via simulated binary crossover and polynomial mutation. From the combined set of parents and the offspring, the worst ranked solution from the largest cluster is eliminated during environmental selection. The performance of the proposed algorithm~(MOMO) is quantitatively assessed using IGDX, PSP and IGD metrics and compared with six other state of the art MMOP algorithms, including MO\_Ring\_PSO\_SCD, DN-NSGAII, TriMOEA-TA\&R, CPDEA, MMOEA/DC and MMEA-WI. The results indicate that MOMO offers significantly better performance than the above listed algorithms based on $21$ test problems from commonly used test suites. We hope that this study would encourage further development of simple and efficient algorithms to solve MMOPs with limited budget, aimed towards a wider uptake in solving practical applications.

A few other interesting directions to pursue include (a) means to handle constraints (b) design of new MMOPs involving constraints and (c) development of surrogate-assisted optimization algorithms which use even fewer function evaluations to deal with such classes of problems. Some of these directions are currently being pursued by the authors.

\balance
\bibliographystyle{IEEEtran}
\bibliography{refs}
\end{document}